\documentclass{article}

\usepackage{microtype}
\usepackage{graphicx}
\usepackage{subfigure}
\usepackage{booktabs} 

\usepackage{hyperref}

\usepackage[accepted]{icml2024}

\usepackage{amsmath}
\usepackage{amssymb}
\usepackage{mathtools}
\usepackage{amsthm}

\usepackage[capitalize,noabbrev]{cleveref}

\theoremstyle{plain}

\theoremstyle{definition}

\theoremstyle{remark}

\usepackage[textsize=tiny]{todonotes}

\usepackage[utf8]{inputenc} 
\usepackage[T1]{fontenc}    
\usepackage{hyperref}       
\usepackage{url}            
\usepackage{booktabs}       
\usepackage{amsfonts}       
\usepackage{nicefrac}       
\usepackage{microtype}      
\usepackage{xcolor}         
\usepackage{wrapfig}
\usepackage{pifont}
\usepackage{amsmath}
\usepackage{amssymb}
\usepackage{mathtools}
\usepackage{amsthm}
\usepackage{multirow}
\usepackage{bm}
\usepackage{enumitem}
\usepackage{tikz}
\usepackage{varwidth}
\icmltitlerunning{Rethinking Independent Cross-Entropy Loss For Graph-Structured Data}

\begin{document}

\twocolumn[
\icmltitle{Rethinking Independent Cross-Entropy Loss For Graph-Structured Data}




\icmlsetsymbol{equal}{*}
\begin{icmlauthorlist}
\icmlauthor{Rui Miao}{1}
\icmlauthor{Kaixiong Zhou}{2}
\icmlauthor{Yili Wang}{1}
\icmlauthor{Ninghao Liu}{3}
\icmlauthor{Ying Wang}{4}
\icmlauthor{Xin Wang}{1}*
\end{icmlauthorlist}

\icmlaffiliation{1}{School of Artificial Intelligence, Jilin University, China.}
\icmlaffiliation{2}{Institute for Medical Engineering Science, Massachusetts Institute of Technology, USA.}
\icmlaffiliation{3}{School of Computing, University of Georgia, USA.}
\icmlaffiliation{4}{College of Computer Science and Technology, Jilin University, China}

\icmlcorrespondingauthor{Xin Wang}{xinwang@jlu.edu.cn}

\icmlkeywords{Machine Learning, ICML}

\vskip 0.3in
]



\printAffiliationsAndNotice{}  

\begin{abstract}
Graph neural networks (GNNs) have exhibited prominent performance in learning graph-structured data. Considering node classification task, based on the i.i.d assumption among node labels, the traditional supervised learning simply sums up cross-entropy losses of the independent training nodes and applies the average loss to optimize GNNs' weights. But different from other data formats, the nodes are naturally connected. It is found that the independent distribution modeling of node labels restricts GNNs' capability to generalize over the entire graph and defend adversarial attacks. In this work, we propose a new framework, termed joint-cluster supervised learning, to model the joint distribution of each node with its corresponding cluster. We learn the joint distribution of node and cluster labels conditioned on their representations, and train GNNs with the obtained joint loss. In this way, the data-label reference signals extracted from the local cluster explicitly strengthen the discrimination ability on the target node. The extensive experiments demonstrate that our joint-cluster supervised learning can effectively bolster GNNs' node classification accuracy. Furthermore, being benefited from the reference signals which may be free from spiteful interference, our learning paradigm significantly protects the node classification from being affected by the adversarial attack.
\end{abstract}

\section{Introduction}

Graph-structured data is ubiquitous in a broad spectrum of application domains, such as social networks~\citep{social1, social2}, biological networks~\citep{bio1, graphsam, PGRGOOD}, recommender system~\citep{recom1, recom3}. Graph neural networks (GNNs) have been extensively explored to learn the complex connectivity information and node features in an end-to-end manner. Particularly, GNNs follow a message passing strategy and learn the representation of each node by iteratively aggregating the representations of its neighbors and combining with itself, which facilitate various downstream tasks including node classification~\citep{GCN, node1}, link prediction~\citep{link1}, and graph classification~\citep{graphc2, graphc1}. 

Despite the persistent efforts in feature learning, label dependencies among nodes receives inadequate attentions. Considering the node classification task with GNNs, decision making is modeled by independent conditional distribution $P(y_i | z_i)$, where $y_i$ and $z_i$ are the label and learned feature of a specific node and its cross-entropy loss is $\mathrm{CE}(y_i, P(y_i|z_i))$. However, it is notorious that such independent decision making of node label exacerbates following issues. \underline{\textit{Overfitting}}: The overly minimization of cross-entropy loss prefers the  higher prediction probabilities (i.e., over-confident decision) on the small set of training nodes, resulting in poor generalization on the rest of graph\citep{overfit1, overcon}. \underline{\textit{Susceptibility to adversarial attacks}}: The over-confident GNNs underestimate their uncertainties, which is often leveraged by adversarial attacker to craft input examples that lie in uncertain regions but have different labels~\citep{adver}. This presents a challenge to calibrate GNNs' training and hence generate robust decision making. 

On the other hand, the decision making based on i.i.d assumption of node label is not in line with the graph-structured data, where nodes tend to connect with ``similar'' neighbors to form some clusters. The i.i.d assumption  factorizes the joint distribution 
into a product of multiple prediction densities: $P(y_1, \cdots, y_n | z_1, \cdots, z_n ) = \prod P(y_i | z_i)$. 
This straightforward factorization fails to comprehensively account for the inherent node correlations. Although the message passing learns the neighborhood-aware node features, the label dependencies are thrown out during node inference. 
Just like human experts making decisions with other data-label pairs as reference signals, GNN models could promote their reasoning capabilities via the prompt data. We are aware of the previous arts in investigating the label dependencies~\citep{cgnf, cs}; but they either cannot unite with the feature learning in GNNs or have poor efficiency. In view of such, we ask: 

\colorbox{lightgray}{\begin{varwidth}{\linewidth}\textit{How can we efficiently learn the joint distribution together with GNNs to reason node label rationally?}\end{varwidth}}

In this paper, we propose a new framework named joint-cluster supervised learning to model the joint distribution of each node and its corresponding cluster. For an individual node, we learn the joint-cluster distribution of $P(y_i, y_c | z_i, z_c)$, where $z_c$ and $y_c$ are the constructed cluster feature and label, respectively. The motivation for adopting cluster is to provide the sufficient reference signals for target sample, while reducing computational complexity required to integrate the remaining nodes in the vanilla joint distribution. Particularly, we optimize GNNs by minimizing the joint-cluster cross-entropy loss. The well-trained GNNs are then leveraged to infer the node label by marginalizing the joint-cluster distribution as shown in Figure~\ref{F3}. \textit{Compared to supervised learning, the main difference of our work is to explicitly learn the joint density of the target sample and its reference signals.}   The contributions are summarized below:

\begin{itemize}[leftmargin=0.4cm, itemindent=.0cm, itemsep=0.0cm, topsep=0.0cm]

    \item We introduce a new paradigm of joint-cluster supervised learning for graph data. By breaking the i.i.d assumption in node classes and loss computation, we propose to model the joint distribution between the target node and its located cluster, and leverage it to train and infer GNNs. 
    
    \item The joint distribution disperses prediction densities of a node over a larger label space, thereby relieving the over-confident decision making. We comprehensively test on small, large, class-imbalanced, and heterophilic graphs. The experiments on 12 datasets and 7 backbone models consistently validate the substantial generalization capability of joint-cluster distribution learning. 
    
    \item The joint-cluster distribution learning generates more robust classifications for the attacked nodes compared with the independent decision making, owing to the reliable reference signal of cluster. 
    
    \item The joint-cluster supervised learning surpasses the state-of-the-art (SOTA) models that encode the label dependencies, in terms of the node classification accuracy, training and inference efficiencies. 
\end{itemize}

\section{Preliminary of GNNs and Supervised Learning}
We focus on the node classification task to introduce the new concept of joint distribution learning. Let $\left(x_{i},y_{i}\right)$ denote the node-label pair, where $x_{i}\in \mathbb{R}^{d}$ and $y_{i}\in \mathbb{R}^{c}$ are input feature and label vector of node $v_i$, resepctively, $d$ and $c$ are the dimension sizes. Given training set $\left\{\left(x_{i},y_{i}\right)\right\}_{i=1}^{L}$, where $L$ is the number of labeled nodes, the goal of node classification task is to train a predictor $f_\theta: \mathbb{R}^{d}\rightarrow\mathbb{R}^{c}$, mapping each node over the entire graph to a desired label with trainable parameter $\theta$.

\subsection{Graph Neural Networks}
\label{sec:gnn}
GNNs have emerged as one of the standard tools to learn both the node features and graph structure. Mathematically, based on the recursive message passing mechanism, at the $k$-th layer of GNNs, the embedding vector $z^{(k)}_i$ of each node $v_i$ is obtained by~\citep{gin}:
\begin{equation}
\small
    \label{eq: gnn}
    z^{(k)}_i = \mathrm{Aggregate}(\{z^{(k-1)}_j \mid \forall j\in \mathcal{N}(i)\cup i\}; \theta).
\end{equation}
Function $\mathrm{Aggregate}$ denotes combination operator (e.g., sum, mean, or max) on the neighborhood embeddings, and $\mathcal{N}(i)$ denotes a set of neighbors connected to node $v_i$. Suppose we have a number of repeated message-passing layers. We simply use $z_i = f_\theta(x_i)$ to denote the final generated node representation of node $v_i$ and utilize it to predict the corresponding node label $\hat{y}_i$.  

\begin{figure*}
\centering
\includegraphics[width=0.8\textwidth]{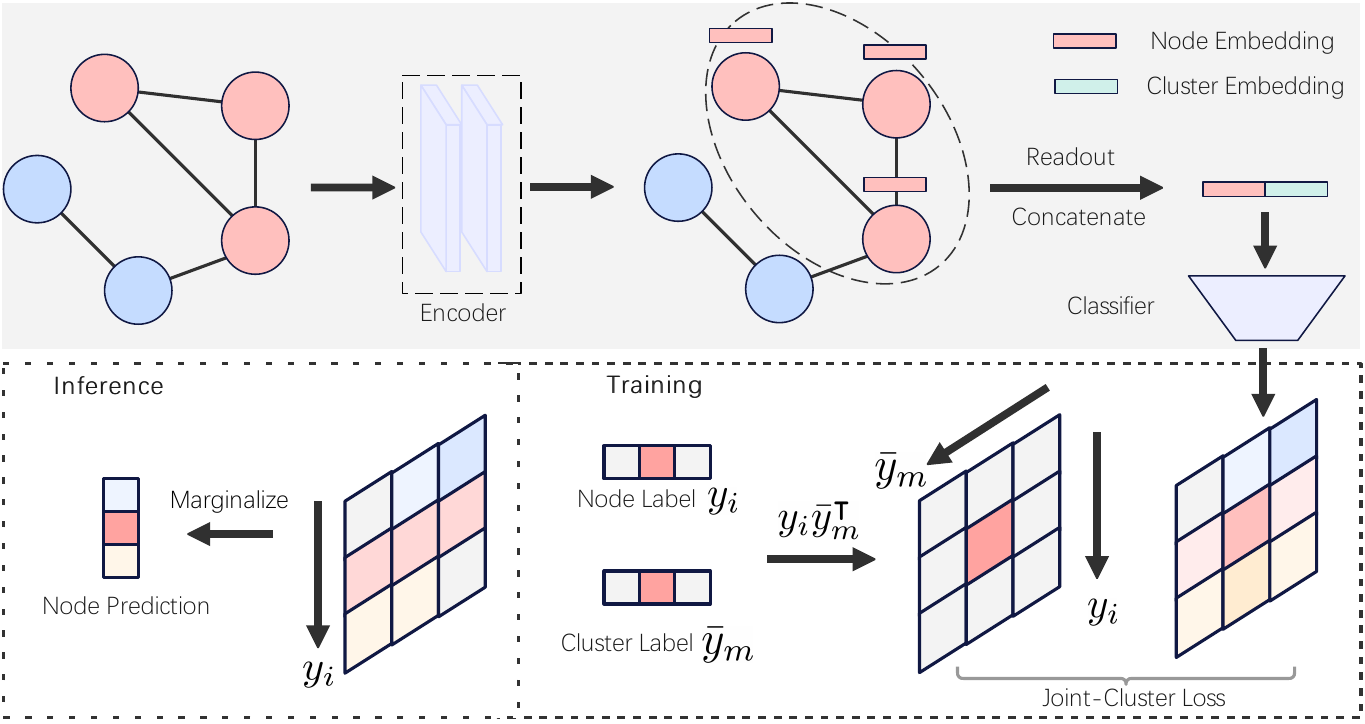}
\vspace{-10pt}
\caption{An illustration of our joint-cluster supervised learning framework: First, we obtain node embeddings through the encoder. Then the cluster embedding and label are generated through the divided graph structure. Then the node embedding and the cluster embedding are concatenated and fed into the classifier to obtain a joint distribution prediction. Finally, the joint-cluster loss and marginalization are used for training and inference.}
\label{F3}
\vspace{-10pt}
\end{figure*}

\subsection{Independent Cross-entropy Loss}
Following the supervised learning paradigm and considering the training nodes, vanilla cross-entropy loss is obtained by $\mathcal{L}_{CE} = -\sum_{i=1}^{L}y_i\log\hat{y}_{i}$. This approach has been applied to multiple domains such as CV and NLP.  
Particularly, the supervised learning makes use of the conditional density $p\left(y\mid z\right)$ for each pair $\left(z_i,y_i\right)$ and train model weights via  maximum likelihood estimator (MLE):
\begin{subequations}
\small
\label{E2}
\begin{align}
 \hat{\theta}_{\mathrm{CE}}\left(\left\{z_{i}, y_{i}\right\}_{i=1}^{L}\right)&\label{E2-1}=\underset{\theta}{\arg \max}  \ p\left(y_{1}, \ldots, y_{L} \mid z_{1}, \ldots, z_{L} ;\theta\right) \\ 
 \label{E2-2}&=\underset{\theta}{\arg \max } \prod_{i=1}^{L} p\left(y_{i} \mid z_{i} ; \theta\right) \\
 \label{E2-3}&=\underset{\theta}{\arg \max } \sum_{i=1}^{L} \log p\left(y_{i} \mid z_{i} ; \theta\right).
\end{align}
\end{subequations}
Note that we use $p\left(y_{i} \mid z_{i}; \theta\right)$ and $p\left(y_{i} \mid z_{i}\right)$ interactively, where the former one is used in the context of model optimization and the later one is adopted for simpleness. The transition from Eq.(\ref{E2-1}) to Eq.(\ref{E2-2}) is deduced not according to mathematical consequence but based on the i.i.d assumption between nodes' labels. However, such decomposition is not desired in graph data, since the node features and classes are inherently correlated depending on the graph connectivity. Although GNNs aggregate the neighborhoods and make decision on the target node conditioned on the set of neighbors' features, the joint-distribution modeling of node classes is still broken in Eq.(\ref{E2-2}). 
The prior knowledge that the nodes at the same cluster share similar labels is widely accepted in many real-world graphs, like social networks. 
These intuitions inspire us to learn the joint distribution of node classes conditioned on their features. 



\vspace{-5pt}
\section{Joint-cluster Supervised Learning}
As analyzed before, GNNs utilize graph structure through the unique message passing, while still treating the node labels as independent from each other during the loss optimization. Despite the conceptual simpleness, it is not trivial to model the joint distribution. Given a number of training nodes, the fully joint conditional distribution $p\left(y_{1}, \ldots, y_{L} \mid z_{1}, \ldots, z_{L}; \theta \right)$ has to be constructed over label state space of $\mathbb{R}^{c^L}$. The optimization on such high-dimensional state space is computationally intractable and hard to generalize on the test nodes. To enable the computation on common hardware, we propose to learn the joint distribution from cluster perspective. It is generally assumed nodes within the same cluster are highly connected while the edge connections between clusters are sparse. 
We thereby divide the graph into $M$ independent clusters $\left\{ \mathcal{C}_{1}, \ldots, \mathcal{C}_{M} \right\}$ and factorize the joint distribution as:
\begin{subequations}\small
\label{E4}
\begin{flalign}
&p\left(y_{1}, \ldots, y_{L} \mid z_{1}, \ldots, z_{L} ; \theta\right) &\\& \quad =\prod_{m=1}^{M} p\left( \left\{ y_{i} \mid v_i \in \mathcal{C}_{m} \right\} \mid \left\{ z_{i} \mid  v_i \in \mathcal{C}_{m} \right\};\theta\right).&
\end{flalign}
\end{subequations}
In other words, the node label distributions between clusters are close to be independent. Although the i.i.d assumption on clusters reduces the computation complexity to some extent, the joint modeling on a subset of nodes is still impractical and is unfriendly to be adopted to infer the classes of test nodes. In this work, for each node representation-label pair $\left(z_{i}, y_{i}\right)$, we instead learn a joint conditional distribution $p\left(y_{i},\bar{y}_{m} \mid z_{i}, \bar{z}_{m}; \theta\right)$, where $\bar{z}_{m}$ and $\bar{y}_{m}$ denote the statistical cluster feature and label, respectively. One of the simplest ways to construct the cluster feature and label is to average the node representations and labels from the training samples within the corresponding cluster, which is adopted in our method. The more advanced solution, like differentiable features and label vectors, could be used to learn the cluster statistics. Given the set of training nodes, MLE optimizes the joint-cluster conditional distribution as:
\begin{subequations}
\small
\begin{flalign}
&\hat{\theta}_{JC}\left(\left\{z_{i}, y_{i}\right\}_{i=1}^{L}\right) &\\
&\quad =\underset{\theta}{\arg \max }  \ p\left(y_{1}, \ldots, y_{L} \mid z_{1}, \ldots, z_{L} ; \theta\right) \label{5a} &\\ 
&\quad =\underset{\theta}{\arg \max } \prod_{m=1}^{M} p\left( \left\{ y_{i} \mid v_i \in \mathcal{C}_{m} \right\} \mid \left\{ z_{i} \mid  v_i \in \mathcal{C}_{m} \right\};\theta\right) \label{5b} &\\
&\quad =\underset{\theta}{\arg \max } \prod_{m=1}^{M} \prod_{i=1}^{\left|\mathcal{C}_{m}\right|} p\left(y_{i}, \bar{y}_{m} \mid z_{i}, \bar{z}_{m} ; \theta\right) \label{5c} &\\
&\quad =\underset{\theta}{\arg \max } \sum_{m=1}^{M} \sum_{i=1}^{\left|\mathcal{C}_{m}\right|} \log p\left(y_{i}, \bar{y}_{m} \mid z_{i}, \bar{z}_{m} ; \theta\right).&
\end{flalign}
\end{subequations}
Through the transition from Eq.(\ref{5b}) to Eq.(\ref{5c}), we decouple the distributions of connected nodes to facilitate computation but still keep the node-cluster relation to realize joint modeling. In this way, we can easily train on a set of individual nodes and extend the well-trained model to estimate the joint distribution of test samples. In this work, we use the graph clustering algorithm of METIS~\citep{metis}, which aims to construct the vertex partitions such that within clusters links are much more than between-cluster links to better capture the community structure of the graph. This partitioning manner is in line with our i.i.d assumption on clusters, where the between-cluster dependencies are negligible. Based on the above joint modeling, we introduce how to train models and infer node classes, and put the pseudo-code in Appendix~\ref{appendix_algori} for further detailed information.

\paragraph{Training with joint-cluster loss.}
We design the joint-cluster cross-entropy loss to learn the node-cluster distribution. Let 
$f_\theta: \left(x_{i}, \bar{x}_{m}\right)\rightarrow \left(y_{i}, \bar{y}_{m}\right)$ be a model to map the node and cluster features  into their corresponding joint label $y_{i}\bar{y}_{m}^\intercal \in \mathbb{R}^{c \times c}$. As shown in Figure~\ref{F3}, the model consists of an encoder (e.g., GNNs) to generate node representations and a classifier to predict the joint label. Recalling Section~\ref{sec:gnn}, the final representation of node $v_i$ is given by $z_i$. We adopt average pooling to define the cluster representation $\bar{z}_{m} = 1/L_{m}\sum_{i=1}^{L_{m}}z_{i}$ and the cluster label $\bar{y}_{m} = 1/L_{m}\sum_{i=1}^{L_{m}}y_{i}$, where $L_{m}$ is the number of labeled nodes within cluster $\mathcal{C}_m$. We then concatenate the node and cluster representations as the joint feature, which is fed into the classifier to predict joint label $y_{i}\bar{y}_{m}^\intercal$. Mathematically, the joint-cluster cross-entropy loss is defined as:
\begin{equation}\small
\label{E8}
\begin{aligned}
&\mathcal{L}_{JC} = -\sum_{i=1}^{L} \{\left(y_{i}\bar{y}_{m}^\intercal\right) \cdot \log g_\phi\left(\text{con}\left(z_{i}, \bar{z}_{m}\right)\right)\\
& + \left(\bar{y}_{m}y_{i}^{\intercal}\right) \cdot \log g_\phi\left(\text{con}\left(\bar{z}_{m}, z_{i}\right)\right)\}.
\end{aligned}
\end{equation}
where $\text{con}\left(\,\cdot,\cdot \right)$ is a vector concatenation operation ordered by the node embedding and its cluster embedding, $g_\phi$ is the classifier, and node $v_{i}$ belongs to cluster $\mathcal{C}_{m}$. The dot product and $\log$ function operate element-wisely. Notably, for the purpose of symmetric joint distribution modeling, at the second item of the above equation, we exchange the position of node and cluster embeddings to predict their label $\bar{y}_{m}y_{i}^{\intercal}$ (i.e., the transpose of $y_{i}\bar{y}_{m}^\intercal$).




\paragraph{Node class inference in joint distribution.}
Based on the joint distribution $p\left(y_{i}, \bar{y}_{m} \mid z_{i}, \bar{z}_{m}; \theta\right)$ between the node and its cluster, we aim to infer every individual node classes as in the standard supervised learning framework. In other words, we have to recover the independent conditional distribution $p(y_i\mid z_i; \theta)$ and make a decision over the label state space $\mathbb{R}^{c}$. The direct solution is to marginalize the joint label along the cluster label dimension: 
\begin{equation}
\label{E7}
\begin{aligned}
p(y_i \mid z_i ; \theta) &= \int_{\mathbb{R}^{d}} \sum_{k=1}^{c} p\left(y_i, \bar{y}_{m}=k \mid z_i, \bar{z} ; \theta\right) q\left(\bar{z}\right) d \bar{z}\\ & \approx \sum_{k=1}^{c} p\left(y_i, \bar{y}_{m}=k \mid z_i, \bar{z}_{m} ; \theta\right).
\end{aligned}
\end{equation}
$q\left(\bar{z}\right)$ denotes the continuous distribution of cluster representation. In practice, since the node is nearly independent to the other clusters, the approximation deduction in Eq.(\ref{E7}) only uses the dwelling cluster feature to obtain the marginalized distribution. As illustrated in Figure~\ref{F3}, given the two-dimensional prediction $p\left(y_{i}, \bar{y}_{m} \mid z_{i}, \bar{z}_{m} ; \theta\right)$ corresponding to truth $y_{i}\bar{y}_{m}^{\intercal}$, we sum the prediction scores row wisely to estimate $p(y_i \mid z_i ; \theta)$. Unlike the standard supervised learning, during model inference, we make use of the cluster reference signal to reason the node classes rationally and robustly. This merit is functionally similar to the in-context learning strategy explored recently~\citep{intext}, where a set of data-label pairs are concatenated with input to guide the language model to make more accurate decisions. As empirically studies in Appendix~\ref{appendix_ic}, in graph-structured data, we observe the joint distribution modeling provides better node classification accuracy compared with the simple concatenation. Furthermore, we provide the more detailed explanations about joint-cluster supervised learning in Appendix~\ref{more_explanations}.

\begin{table*}[t]\small
  \caption{Test Accuracy (\%) for different models on class-balanced small datasets, where the best results are in bold. 
  CE denotes the cross-entropy loss, and JC denotes our joint-cluster loss function.}
  \label{main_res}
  \centering
  \vskip 0.05in
  \setlength{\tabcolsep}{9.5pt}
  \begin{tabular}{cccccccc}
    \toprule
    Model  & Loss  & Cora & CiteSeer & PubMed  & DBLP  & Facebook\\
    \midrule
    \multirow{2}{*}{GCN} & CE     &$81.70_{\pm0.65}$  &$71.43_{\pm0.47}$    &$79.06_{\pm0.32}$   &$74.30_{\pm 1.94}$  &$73.91_{\pm 1.40}$\\
                         & JC    &$\mathbf{83.51_{\pm 0.35}}$  &$\mathbf{72.97_{\pm 0.55}}$    &$\mathbf{79.80_{\pm 0.19}}$  &$\mathbf{75.10_{ \pm 1.63}}$  &$\mathbf{74.64_{\pm 1.75}}$\\
    \hline
     \multirow{2}{*}{SGC} & CE   &$81.68_{\pm0.52}$  &$71.85_{\pm0.39}$    &$78.70_{\pm0.38}$  & $74.30_{\pm 2.12}$ &$ 74.13_{\pm 2.13}$\\
                          & JC  &$\mathbf{83.87_{\pm 0.79}}$  &$\mathbf{72.92_{\pm 0.16}}$    &$\mathbf{79.97_{\pm 0.25}}$   & $\mathbf{74.87_{\pm 1.81}}$ & $\mathbf{74.74_{\pm 1.96}}$   \\
    \hline
    \multirow{2}{*}{SAGE} & CE     &$79.96 _{\pm 0.44}$  &$69.94_{ \pm 0.93}$    &$78.37_{ \pm 0.72}$   &$70.59_{\pm 1.46}$  &$70.95_{\pm 2.26}$\\ 
                         & JC    &$\mathbf{80.81_{\pm 0.63}}$  &$\mathbf{70.54_{\pm 1.49}}$    &$\mathbf{79.50_{\pm 1.02}}$  &$\mathbf{71.87_{ \pm 2.07}}$  &$\mathbf{71.59 _{\pm 1.78}}$\\

    \hline
    \multirow{2}{*}{GAT} & CE    & $83.22_{\pm 0.29}$ & $71.06_{ \pm 0.40}$ & $78.54_{ \pm 0.63}$   & $75.32_{\pm 2.62}$   & $76.34_{\pm 2.26}$\\
                         & JC   & $\mathbf{83.77_{ \pm 0.44}}$ & $\mathbf{71.61 _{\pm 0.95}}$ & $\mathbf{79.35_{ \pm 0.47}}$  & $\mathbf{76.92_{\pm 1.59}}$  & $\mathbf{77.46_{\pm 2.30}}$\\
    \hline
    \multirow{2}{*}{MLP} & CE   & $58.65_{ \pm 0.97}$ & $60.41_{ \pm 0.56}$ & $73.27_{ \pm 0.35}$ & $47.95_{ \pm 3.97}$  & $55.34_{ \pm 2.60}$\\
                        & JC   & $\mathbf{67.19_{ \pm 0.62}}$ & $\mathbf{63.23_{ \pm 0.87}}$ & $\mathbf{75.92_{\pm 0.39}}$  & $\mathbf{61.16_{ \pm 3.63}}$  & $\mathbf{56.62_{\pm 2.42}}$\\
  \bottomrule
\end{tabular}
\vspace{-5pt}
\end{table*}

\section{Related Work}
A detailed discussion is provided in Appendix~\ref{appendix_related}. Two families of label dependency modeling are:

\paragraph{Label propagation.} 
In the realm of GNNs, label propagation works on the assumption that nodes connected by an edge are likely to share the same label~\citep{shi2020masked, wang2020unifying, zhou_ALS}. It propagates node labels along with edge weights throughout the graph~\citep{wang2021combining, xie2022graphhop}, and then infer the unlabeled nodes effectively.
\textit{\underline{Difference compared to existing work}}: While the label propagation often infer nodes without considering the node features (e.g., at post-processing phase), our joint-cluster learning framework could work with any GNN backbones to comprehensively learn the structure, feature, and label information end-to-end.

\paragraph{Conditional random fields.}
To leverage the label correlation in node classification, there has been previous work in combining conditional random fields (CRF) with GNNs. CGNF~\citep{cgnf} learns the pairwise label correlation with pairwise energy function, a specific expression form of CRF, which is optimized to train GNNs. CRF-GNNs inserts CRF layer between the graph convolutional layers, which regularizes GNNs to preserve the label dependencies among nodes~\citep{crf-gnn}. SPN~\citep{spn} models the local label correlation of each linked node pair via nodeGNN and edgeGNN and takes the edges in graph as input to propagate all the pairwise label correlations along edges. \textit{\underline{Difference compared to existing work}}: \ding{172} While CRF-based methods focus on modeling the local label correlation of every linked node pair, we learn the global joint distribution of node and its cluster. \ding{173} Our proposals show promising training scalability and inference efficiency. The CRF-based methods take the whole graph as input to propagate all the pairwise label correlations along edges. In contrast, we train and infer the joint distribution of target node only with one reference signal (i.e., cluster), which allows the batch training on large graphs (e.g., Amazon with millions of nodes). 

\section{Experiments}
In this section, we evaluate our joint-cluster learning framework on 12 public datasets over 7 backbone models. The code is available at: \href{https://github.com/MR9812/Joint-Cluster-Supervised-Learning}{https://github.com/MR9812/Joint-Cluster-Supervised-Learning}


\vspace{-0.1cm}
\subsection{Evaluation on Small Graph Datasets}
\vspace{-0.2cm}

\paragraph{Implementation.} $\rhd$ \textsf{Datasets}. We use the benchmark datasets Cora, CiteSeer, PubMed~\citep{citation_network}, DBLP~\citep{DBLP}, and Facebook~\citep{Facebook} in the class-balanced setting, which is widely adopted to evaluate GNNs. Furthermore, we consider two more challenging node classification tasks. In particular, we conduct on LastFMAsia~\citep{LastFM} and ogbn-arxiv~\citep{ogb} to evaluate the performance of our proposed joint-cluster loss on class-imbalanced environment; and we consider Chameleon, Squirrel, and Wisconsin to evaluate on heterophilic graphs~\citep{cham}. $\rhd$ \textsf{Backbone models}. We take GCN~\citep{GCN}, SGC~\citep{SGC}, GraphSage~\citep{GraphSage}, GAT~\citep{GAT} and MLP as base models to compare our proposals with the standard cross-entropy loss in the class-balanced setting. Due to space limit, we use GCN, SGC, and MLP to evaluate on the class-imbalanced and heterophilic environment. 
The details of datasets and backbone models are presented in Appendix~\ref{appendix_data}~\ref{appendix_backbone}. We run each experiment 10 times and report the mean values with standard deviation. 

\vspace{-0.2cm}
\paragraph{Q: Whether our proposals outperform the standard supervised learning on the easy and small datasets?} Yes, one key advantage of joint distribution modeling is to infer nodes more correctly with cluster references. We examine on class-balanced, class-imbalanced, and heterophilic datasets.


$\rhd$ \textsf{Class-balanced graph datasets}. The comparison results are collected in Table\ref{main_res}, from which we make following observations. \ding{182} \textit{The joint-cluster supervised learning exhibits significantly superior performances on all the backbone models.}
Compared with the standard cross-entropy loss, our approach delivers the average improvements of $1.47\%$, $1.47\%$, $1.21\%$, $1.22\%$ on models GCN, SGC, SAGE, and GAT, respectively. 
\ding{183} Interestingly, compared with the average improvement of 1.34\% over GNN backbones, the more clear advantage of 10.5\% is achieved in MLP architecture. That is because GNNs learn the single  node class conditioned on aggregated features, while MLP decides the node label only based on its input feature. Moving a step forward, our proposals learn the comprehensive joint distribution of multiple node labels conditioned on their features aggregated from GNNs, which fully activates the model's generalization ability.


\begin{table*}\small
  \centering
  \setlength{\tabcolsep}{5.0pt}
  \caption{Test F1(\%) of  different loss functions on the class-imbalanced datasets.} 
  \label{imbalance_res}
  \vskip 0.05in
  \begin{tabular}{cccccccc}
   \toprule
   \multirow{2}{*}{Model} & \multirow{2}{*}{Loss} & \multicolumn{3}{c}{LastFMAsia} & \multicolumn{3}{c}{ogbn-arxiv} \\
   \cmidrule{3-8} &    & F1-micro  & F1-macro  & F1-weight   & F1-micro  & F1-macro   & F1-weight  \\
   \hline
   \multirow{2}{*}{GCN} & CE     &$84.91_{\pm 0.74}$  &$73.79_{\pm 1.28}$   &$84.60_{\pm 0.73}$  &$71.74_{\pm 0.29}$   &$51.80_{\pm 0.44}$  &$70.93_{\pm 0.33}$ \\
                         & JC    &$\mathbf{85.92_{\pm 0.41}}$  &$\mathbf{74.61_{\pm 1.02}}$  &$\mathbf{85.49 _{\pm 0.43}}$  &$\mathbf{72.17_{ \pm 0.24}}$  &$\mathbf{52.06 _{\pm 0.15}}$  &$\mathbf{71.57_{ \pm 0.18}}$\\
   \hline
   \multirow{2}{*}{SGC} & CE     &$84.82_{\pm 0.82}$  &$70.85_{\pm 1.36}$   &$84.22_{\pm 0.55}$  &$71.77_{\pm 0.14}$   &$50.75_{\pm 0.29}$  &$70.71_{\pm 0.21}$\\
                         & JC    &$\mathbf{85.84_{\pm 0.45}}$  &$\mathbf{73.25_{\pm 1.17}}$  &$\mathbf{85.32_{\pm 0.40}}$  &$\mathbf{72.08_{\pm 0.15}}$   &$\mathbf{51.15 _{\pm 0.26}}$  &$\mathbf{70.92_{ \pm 0.15}}$\\
   \hline
   \multirow{2}{*}{MLP} & CE    &$68.91_{\pm 0.70}$  &$42.37_{\pm 1.45}$   &$67.06_{\pm 0.73}$  & $55.50_{ \pm 0.23}$   &$33.93_{\pm 0.20}$  &$55.00_{\pm 0.19}$\\
                         & JC   &$\mathbf{78.81_{\pm 0.69}}$  &$\mathbf{53.09_{\pm 1.71}}$  &$\mathbf{76.89_{ \pm 0.69}}$ &$\mathbf{61.21_{\pm 0.16}}$  &$\mathbf{38.61 _{\pm 0.23}}$  &$\mathbf{60.48_{ \pm 0.11}}$\\                
   \bottomrule
  \end{tabular}%
  \vspace{-10pt}
 \end{table*}

$\rhd$ \textsf{Class-imbalanced graph datasets}. As shown in Table\ref{imbalance_res}, we observe \ding{182} \textit{the similar trend of  performance enhancement in the imbalance setting.}
We use imbalance ratio, $min_{i}\left(\left|\mathcal{T}_{i}\right|\right)/max_{i}\left(\left|\mathcal{T}_{i}\right|\right)$, to measure the extent of class imbalance, where $\left|\mathcal{T}_{i}\right|$ represents the number of nodes belonging to the $i$-th class. LastFMAsia and ogbn-arxiv are two extremely imbalanced datasets, whose imbalance rates are 1.0\% and 0.1\%, respectively. 
It is observed our joint-cluster learning framework obtains  average improvements of 5.75\% and 3.16\% on LastFMAsia and ogbn-arxiv over the standard supervised learning. We attribute this result to the referential ability of the joint-cluster distribution modeling, which uses the cluster of neighbors when making decisions. The joint distribution weakens the over-confident prediction on the majority classes by assigning prediction confidence on other related minority classes, and thus ameliorates the generalization on them.

$\rhd$ \textsf{Heterophilic graph datasets}. On the heterophilic graphs, the connected nodes tend to have the different classes and make the joint-distribution learning challenging via adding label noise. Following the data split of~\citet{geom-gcn}, we compare with vanilla cross-entropy loss on three benchmark datasets. As shown in Table~\ref{tab: hete}, \ding{182} \textit{we observe our joint-cluster loss function consistently delivers great advantage with clear performance margin.} That is because the proposed joint-cluster distribution learning infer node label with the reference signal of whole cluster, instead of using the direct neighbors. This validates the effectiveness of adopting global cluster structure in joint distribution. 

\begin{table}[t]
\caption{Test accuracy (\%) on heterophilic graphs.}
\label{tab: hete}
\vskip 0.05in
\setlength{\tabcolsep}{1.5pt}
\begin{center}
\begin{small}
\begin{tabular}{ccccc}
\toprule
    Model  & Loss & Chameleon & Squirrel &  Wisconsin \\
    \midrule
    \multirow{2}{*}{GCN} & CE & $59.25_{\pm 2.81}$ &$48.93_{\pm 2.21}$ &$49.22_{\pm 3.77}$ \\ 
& JC & $\mathbf{68.87_{\pm 2.55}}$ &$\mathbf{56.76_{\pm 1.43}}$ &$\mathbf{50.39_{\pm 4.53}}$ \\ 
\hline
\multirow{2}{*}{SGC}  & CE &$63.88 _{\pm 2.78}$ &$53.79 _{\pm 3.13}$ &$51.96 _{\pm 4.23}$ \\
& JC &$\mathbf{71.91_{ \pm 2.03}}$ &$\mathbf{61.99_{ \pm 2.42}}$ &$\mathbf{52.23_{ \pm 3.94}}$ \\
\hline
\multirow{2}{*}{MLP}  & CE &$41.90_{\pm 1.51}$ &$29.23_{\pm 2.09}$ &$80.98_{\pm 5.12}$ \\
& JC &$\mathbf{50.09_{ \pm 2.42}}$ &$\mathbf{32.37_{ \pm 2.20}}$ &$\mathbf{81.96_{ \pm 5.45}}$ \\
\bottomrule
\end{tabular}
\end{small}
\end{center}
\vspace{-10pt}
\end{table}


\subsection{Evaluation on Large Graph Datasets}

\paragraph{Implementation.}
$\rhd$ \textsf{Datasets}. Two complex large datasets are adopted, i.e., Yelp and Amazon~\citep{graphsaint}, where each node contains multiple classes. $\rhd$ \textsf{Backbone models}. We evaluate in two scalable sub-graph sampling models, i.e., GraphSAGE~\citep{GraphSage} and Cluster-GCN~\citep{cluster_gcn}, and in pre-computing-based model of SIGN~\citep{sign}.
The details of datasets and backbone models are provided in Appendix~\ref{appendix_data}~\ref{appendix_backbone}.



\paragraph{Q: Whether our proposals can scale on the large datasets and boost model performance?} Yes, as reported in Table \ref{multi-label}, \ding{182} \textit{the joint-cluster learning framework generally obtains the best accuracy on the large-scale multi-class datasets.} 
Compared with the standard cross-entropy, 
our method obtains the average improvement of 0.75\% and 0.39\% on Yelp and Amazon, respectively. 
One exceptional cases is SIGN conducting on Amazon dataset. We speculate that one of the main reasons is the batch size, which is not large enough to obtain enough cluster statistics for the joint-cluster distribution modeling. The future work can use the trainable cluster feature and label to overcome this problem. 

\begin{table}[t]
\caption{Test micro-F1(\%) on large graph datasets.}
\label{multi-label}
\vskip 0.05in
\setlength{\tabcolsep}{1.5pt}
\begin{center}
\begin{small}
\begin{tabular}{cccc}
\toprule
    Model  & Loss  & Yelp & Amazon \\
    \midrule
    \multirow{2}{*}{GraphSAGE} & CE               & $63.67_{\pm 0.38}$ & $75.65_{\pm 0.16}$\\
    & JC               & $\mathbf{63.99_{\pm 0.46}}$ & $\mathbf{76.14_{\pm 0.29}}$ \\

    \hline
    \multirow{2}{*}{Cluster-GCN} & CE               & $62.44_{\pm 0.52}$ & $76.12_{\pm 0.17}$\\
    & JC               & $\mathbf{63.02_{\pm 0.68}}$ & $\mathbf{76.63_{\pm 0.27}}$ \\

    \hline
     \multirow{2}{*}{SIGN} & CE               & $64.42_{\pm 0.07}$ & $\mathbf{80.22_{\pm 0.04}}$ \\
    & JC               & $\mathbf{64.95_{\pm 0.09}}$ & $80.09_{\pm 0.05}$ \\
\bottomrule
\end{tabular}
\end{small}
\end{center}
\vspace{-10pt}
\end{table}

\begin{table*}[t]\small
  \centering
  \setlength{\tabcolsep}{4pt}
  \caption{Test accuracy (\%) under metattack, where Ptb Rate means the perturbation percent.} 
  \vskip 0.05in
  \label{adv_res}
  \begin{tabular}{cccccccc}
   \toprule
   \multirow{2}{*}{Datasets} & \multirow{2}{*}{Ptb Rate(\%)} & \multicolumn{2}{c}{GCN} & \multicolumn{2}{c}{SGC} & \multicolumn{2}{c}{GAT} \\
   \cmidrule{3-8} &    & CE  & JC  & CE   & JC  & CE   & JC  \\
   \hline
   \multirow{5}{*}{Cora}
                         & 5\%    &$76.80_{\pm 0.87}$  &$\mathbf{78.84_{\pm 0.57}}$  &$76.28 _{\pm 0.20}$  &$\mathbf{78.76_{ \pm 0.45}}$  &$80.24_{\pm 0.54}$  &$\mathbf{80.76_{ \pm 0.51}}$\\
                         & 10\%    &$70.12_{\pm 1.42}$  &$\mathbf{74.65_{\pm 0.44}}$  &$69.29 _{\pm 0.48}$  &$\mathbf{73.50_{ \pm 0.57}}$  &$74.89_{\pm 1.46}$  &$\mathbf{75.49_{ \pm 0.83}}$\\
                         & 15\%    &$64.21_{\pm 1.92}$  &$\mathbf{72.24_{\pm 0.67}}$  &$65.05 _{\pm 1.09}$  &$\mathbf{71.93_{ \pm 0.51}}$  &$70.55 _{\pm1.19}$  &$\mathbf{71.63_{ \pm 1.29}}$\\
                         & 20\%    &$53.56_{\pm 1.98}$  &$\mathbf{59.77_{\pm 0.75}}$  &$57.14 _{\pm 0.32}$  &$\mathbf{58.11_{ \pm 0.83}}$  &$58.74 _{\pm 1.60}$  &$\mathbf{59.45_{ \pm 1.12}}$\\
                         & 25\%    &$48.98_{\pm 1.58}$  &$\mathbf{53.89_{\pm 1.19}}$  &$51.18 _{\pm 0.51}$  &$\mathbf{53.44_{ \pm 1.04}}$  &$53.38 _{\pm 1.14}$  &$\mathbf{55.46_{ \pm 1.68}}$\\
   \hline
   \multirow{5}{*}{CiteSeer} 
                         & 5\%    &$69.96_{\pm 0.82}$  &$\mathbf{70.15_{\pm 0.79}}$  &$71.87 _{\pm 0.20}$  &$\mathbf{72.72_{ \pm 0.62}}$  &$72.03 _{\pm 1.08}$  &$\mathbf{73.96_{ \pm 0.53}}$\\
                         & 10\%    &$67.39_{\pm 0.74}$  &$\mathbf{68.51_{\pm 1.06}}$  &$68.19 _{\pm 0.15}$  &$\mathbf{68.84_{ \pm 0.50}}$  &$70.21 _{\pm 0.82}$  &$\mathbf{71.10_{ \pm 0.24}}$\\
                         & 15\%    &$64.32_{\pm 0.93}$  &$\mathbf{67.23_{\pm 1.24}}$  &$65.01 _{\pm 1.68}$  &$\mathbf{67.59_{ \pm 0.79}}$  &$67.99 _{\pm 1.43}$  &$\mathbf{70.39_{ \pm 0.57}}$\\
                         & 20\%    &$55.18_{\pm 1.67}$  &$\mathbf{57.59_{\pm 1.29}}$  &$56.38 _{\pm 0.23}$  &$\mathbf{56.67_{ \pm 0.88}}$  &$60.40 _{\pm 1.41}$  &$\mathbf{61.57_{ \pm 0.98}}$\\
                         & 25\%    &$56.22_{\pm 2.27}$  &$\mathbf{61.54_{\pm 2.01}}$  &$55.94 _{\pm 0.14}$  &$\mathbf{61.75_{ \pm 0.92}}$  &$59.60 _{\pm 2.18}$  &$\mathbf{60.74
                         _{ \pm 1.05}}$\\
   \hline
    \multirow{5}{*}{PubMed} 
                         & 5\%    &$83.09_{\pm 0.10}$  &$\mathbf{83.17_{\pm 0.10}}$  &$78.12 _{\pm 0.03}$  &$\mathbf{83.07_{ \pm 0.07}}$  &$82.27_{\pm 0.19}$  &$\mathbf{82.97_{ \pm 0.30}}$\\
                         & 10\%    &$81.08_{\pm 0.18}$  &$\mathbf{81.27_{\pm 0.10}}$  &$71.16 _{\pm 0.00}$  &$\mathbf{81.35_{ \pm 0.06}}$  &$79.93_{\pm 0.16}$  &$\mathbf{81.81_{ \pm 0.39}}$\\
                         & 15\%    &$78.31_{\pm 0.28}$  &$\mathbf{78.71_{\pm 0.07}}$  &$67.16 _{\pm 0.03}$  &$\mathbf{78.85_{ \pm 0.06}}$  &$78.24 _{\pm 0.13}$  &$\mathbf{80.08_{ \pm 0.24}}$\\
                         & 20\%    &$76.55_{\pm 0.34}$  &$\mathbf{76.90_{\pm 0.08}}$  &$63.88 _{\pm 0.02}$  &$\mathbf{77.03_{ \pm 0.09}}$  &$75.83 _{\pm 0.27}$  &$\mathbf{78.02_{ \pm 0.34}}$\\
                         & 25\%    &$74.51_{\pm 0.50}$  &$\mathbf{75.05_{\pm 0.07}}$  &$61.10 _{\pm 0.01}$  &$\mathbf{75.05_{ \pm 0.11}}$  &$73.01_{\pm 0.35}$  &$\mathbf{75.61_{ \pm 0.32}}$\\
    \hline
   \multirow{5}{*}{Polblogs}
                         & 5\%    &$72.70_{\pm 0.60}$  &$\mathbf{78.37_{\pm 3.57}}$  &$74.44 _{\pm 0.38}$  &$\mathbf{78.70_{ \pm 3.81}}$  &$76.56_{\pm 0.74}$  &$\mathbf{78.65_{ \pm 0.88}}$\\
                         & 10\%    &$71.90_{\pm 0.69}$  &$\mathbf{74.15_{\pm 0.39}}$  &$70.46 _{\pm 0.29}$  &$\mathbf{76.64_{ \pm 0.51}}$  &$72.42_{\pm 0.69}$  &$\mathbf{76.79_{ \pm 1.11}}$\\
                         & 15\%    &$67.92_{\pm 0.78}$  &$\mathbf{70.53_{\pm 0.53}}$  &$55.99 _{\pm 1.85}$  &$\mathbf{72.64_{ \pm 1.24}}$  &$61.13 _{\pm5.36}$  &$\mathbf{69.04_{ \pm 2.97}}$\\
                         & 20\%    &$57.76_{\pm 0.37}$  &$\mathbf{62.84_{\pm 0.93}}$  &$51.94 _{\pm 0.11}$  &$\mathbf{65.60_{ \pm 1.77}}$  &$51.96 _{\pm 0.17}$  &$\mathbf{54.13_{ \pm 0.16}}$\\
                         & 25\%    &$56.17_{\pm 2.11}$  &$\mathbf{64.87_{\pm 0.96}}$  &$52.02 _{\pm 0.46}$  &$\mathbf{61.67_{ \pm 4.12}}$  &$49.46 _{\pm 2.41}$  &$\mathbf{52.04_{ \pm 1.15}}$\\
   \bottomrule
  \end{tabular}%
  \vspace{-5pt}
 \end{table*}

 \begin{table*}\small
  \caption{Accuracy (\%), training time (s), and inference time (s) comparisons with CRF-based models. Since CRF-GCN does not provide code, the accuracy is directly reported and the time is omitted.}
  \label{crf-gcn}
  \vskip 0.05in
  \centering
  \setlength{\tabcolsep}{2pt}
    \begin{tabular}{cccccccccc}
   \toprule
   \multirow{2}{*}{Methods} & \multicolumn{3}{c}{Cora} & \multicolumn{3}{c}{CiteSeer} & \multicolumn{3}{c}{PubMed} \\
   \cmidrule{2-10} & Accuracy  & Training   & Inference & Accuracy   & Training  & Inference & Accuracy & Training  & Inference  \\
    \midrule
    GCN  & $81.70$ & $0.002$ & $0.001$ & $71.43$ & $0.002$ & $0.001$ & $79.06$ & $0.008$ & $0.001$ \\
    CGNF & $83.2$ & $0.389$ & $0.181$ & $72.2$ & $0.240$ & $0.093$ & $79.4$ & $7.523$ & $2.959$ \\
    CRF-GCN  & $82.8$ & $-$ & $-$ & $72.1$ & $-$ & $-$ &$79.2$ & $-$ & $-$ \\
    SPN  & $80.90$ & $0.244$ & $0.162$ & $69.97$ & $0.164$ & $0.121$ &$76.66$ & $2.704$ & $1.254$ \\
    GCN+JC  & $\mathbf{83.51}$ & $0.004$ & $0.001$ & $\mathbf{72.97}$ & $0.005$ & $0.001$ & $\mathbf{79.80}$ & $0.018$ & $0.001$ \\
  \bottomrule
\end{tabular}
\vspace{-5pt}
\end{table*}

\subsection{Robustness Under Adversarial Attack}
\paragraph{Implementation.} Following the previous work, we use datasets including Cora, CiteSeer, PubMed and Polblogs to evaluate robustness under an untargeted adversarial graph attack. Specifically, we use the metattack~\citep{metaattack} implemented in DeepRobust\footnote{https://github.com/DSE-MSU/DeepRobust}, a pytorch library, to generate attacked graphs by deliberately modifying the graph structure. The details are summarized in Appendix~\ref{appendix_data}. Following previous works~\cite{prognn}, we only consider the largest connected component (LCC) in the adversarial graphs, and randomly split 10\%/10\%/80\% of nodes for training, validation, and testing.

\paragraph{Q: Compared with vanilla training, whether the joint-distribution learning can ameliorate model's robustness under adversarial attack?} Yes, the comparison results are collected in Table~\ref{adv_res}, where we make the following observations to support our answers. 
\ding{182} \textit{The joint-cluster learning framework achieves significant gain under all perturbation rates.}
Compared with the independent decision making, our joint-cluster modeling takes the whole cluster as reference signals, which contains certain number of clean nodes to improve the robustness of class prediction. \ding{183} \textit{The performance gain increases with the perturbation rates.} Specifically, the absolute improvements over the vanilla loss are 2.2\%, 3.1\%, 8.6\%, 6.7\% and 8.9\% in the perturbation rates of 5\%, 10\%, 15\%, 20\%, and 25\%. These results validate the effectiveness of cluster reference signal, which is structrually stable even under the acute attacks. 


\subsection{Comparison with Label Dependency Modeling Related Work}
\textbf{Q: Whether the joint-cluster supervised learning delivers the superior accuracy and efficiency compared with existing label dependency learning frameworks?} Yes, we examine it below. 

$\rhd$ \textsf{Comparison with CRF-based models}. We consider threee CRF-based models, i.e., CGNF, CRF-GCN and SPN, and collect the comparison results in Table~\ref{crf-gcn}. \ding{182} \textit{Our proposals obtain the clear performance gains even compared with SOTA models encoding label dependency.} Particularly, the absolute improvements are 0.4\%, 1.1\% and 0.5\% on Cora, Citeseer, and Pubmed, respectively. These baselines predict the target node by accounting the label dependencies from all the connected neighbors. In contrast, we only take the cluster as reference signal to learn the joint distribution, which is simple but shows great generalization. \ding{183} \textit{Our proposals consume much less training and inference times, which are comparable to vanilla GCN.} While we only consider the cluster in joint distribution, the CRF-based models learn the node together with all its neighbors burdensomely.

$\rhd$ \textsf{Concatenating with label propagation}.
C\&S~\citep{cs} is proposed to smooth node labels at the post-processing phase of MLP model. Prior to such post-processing, 
our joint distribution labeling can be plugged in to better prepare MLP by learning the label correlations of nodes. We examine our thoughts in Table\ref{cs}. \ding{182} \textit{It is observed that over all the larger datasets (i.e., except Cora and Citeseer), MLP can evidently benefit from the joint-cluster loss.} 
On the small datasets, the stacking of C\&S and joint loss will make the node labels overly similar over the whole graph and degrade model performance.


\begin{table*}[h]\small
  \caption{Performance of C\&S with the MLP trained by cross-entropy loss and joint-cluster loss.}
  \label{cs}
  \vskip 0.05in
  \centering
  \begin{tabular}{cccccccc}
    \toprule
    Methods    & Cora & CiteSeer & PubMed & DBLP & Facebook & LastFMAsia & Arxiv \\
    \midrule
    MLP+CE       &$58.65_{\pm0.97}$ & $60.41_{\pm0.56}$ & $73.27_{\pm0.35}$ & $47.95_{\pm3.97}$  & $55.34_{\pm2.60}$ & $68.91_{\pm0.70}$ & $55.50_{\pm0.23}$\\
    MLP+JC     &$67.19_{\pm0.62}$ & $63.23_{\pm0.87}$ & $75.92_{\pm0.39}$ & $61.16_{\pm3.63}$ & $56.62_{\pm2.42}$ & $78.81_{\pm0.69}$ & $63.13_{\pm0.10}$\\
    MLP+CE+C\&S     & $\bm{80.05_{\pm0.46}}$ &$\mathbf{70.36_{\pm0.44}}$  & $77.08_{\pm0.26}$ & $71.19_{\pm2.59}$ & $67.48_{\pm4.60}$ & $85.73_{\pm0.61}$ & $68.58_{\pm0.05}$\\
    MLP+JC+C\&S  & $77.37_{\pm0.65}$ & $69.01_{\pm0.93}$ & $\mathbf{77.91_{\pm0.43}}$ 
    & $\mathbf{73.23_{\pm0.86}}$ &$\mathbf{69.39_{\pm3.53}}$ & $\mathbf{87.26_{\pm0.54}}$ & $\mathbf{70.06_{\pm0.09}}$\\
  \bottomrule
\end{tabular}
\vspace{-15pt}
\end{table*}

\subsection{In-depth Discussion of Joint-cluster Supervised Learning}

\begin{figure*}[h]
\centering 
\subfigure { 
\hspace{-15pt}
\includegraphics[width=0.6\columnwidth]{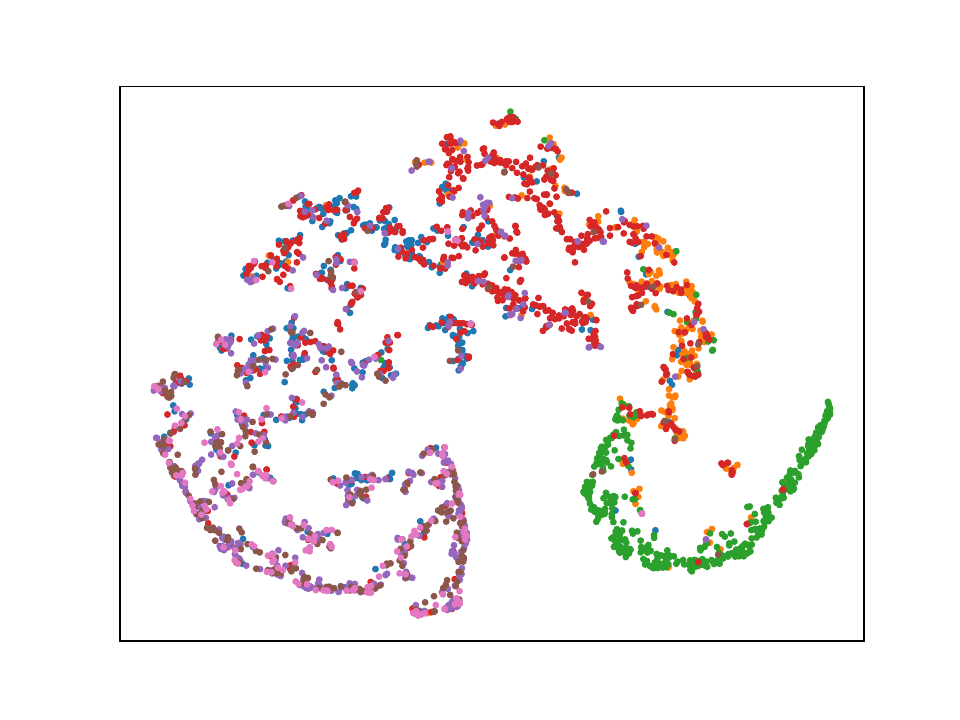} 
}
\subfigure { 
\hspace{-14pt}
\includegraphics[width=0.6\columnwidth]{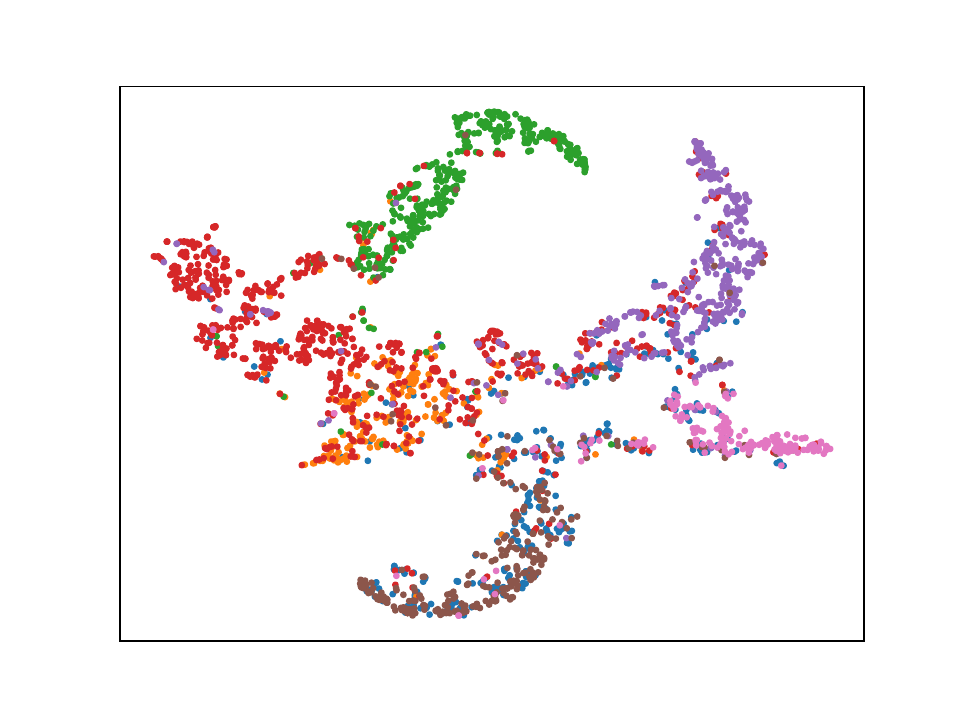} 
}
\raisebox{3pt}{\subfigure { 
\includegraphics[width=0.6\columnwidth]{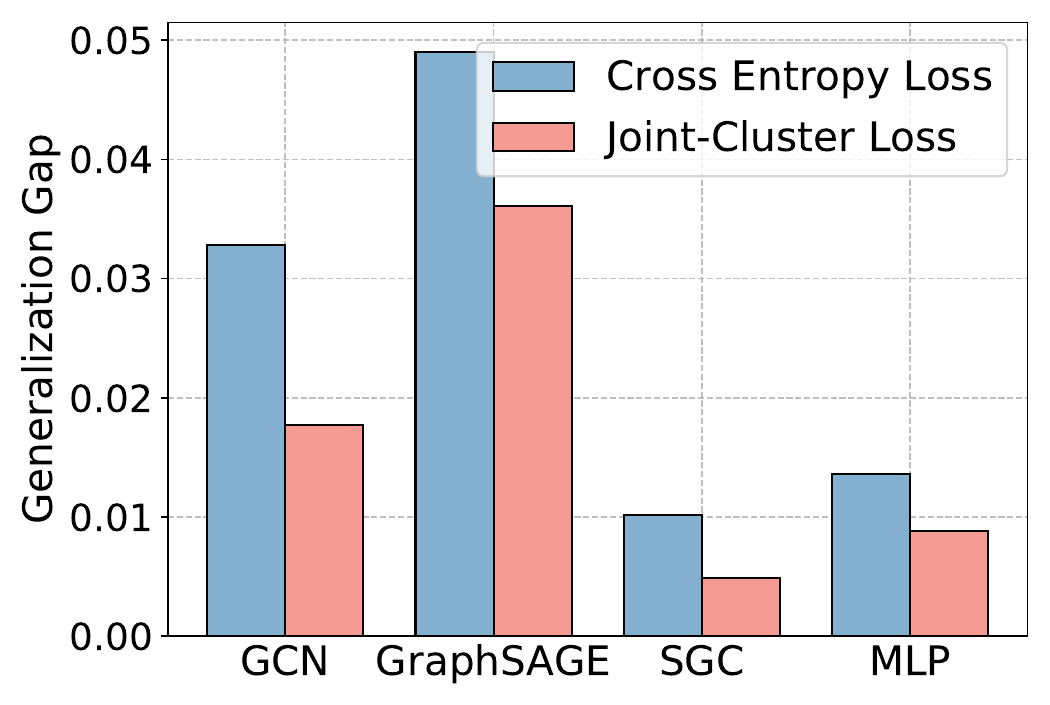}
}}
\vspace{-15pt}
\caption{Left, Middle: Node representation visualization by t-SNE~\citep{tsne} for 8-layer GCN trained by CE loss (left) and JC loss (middle) on Cora. Right: Normalized comparison of the gap between train and test losses on ogbn-arxiv.} 
\label{t-sne}
\vspace{-1pt}
\end{figure*}

\textbf{Q: How the joint-cluster distribution modeling learns to concentrate node embeddings of the same class (cluster) within compact space?}
We visualize the node representations learned by cross-entropy loss and joint-cluster loss in Figure~\ref{t-sne}. 
Different from the vanilla loss, our joint-cluster loss exhibits 2D projections with more coherent shapes of clusters. One of the possible reasons is the node representations are learned to embrace their corresponding clusters in the joint modeling. 

\begin{figure*}[!htb]
\centering 
\vspace{-12pt}
\subfigure { 
\includegraphics[width=0.65\columnwidth]{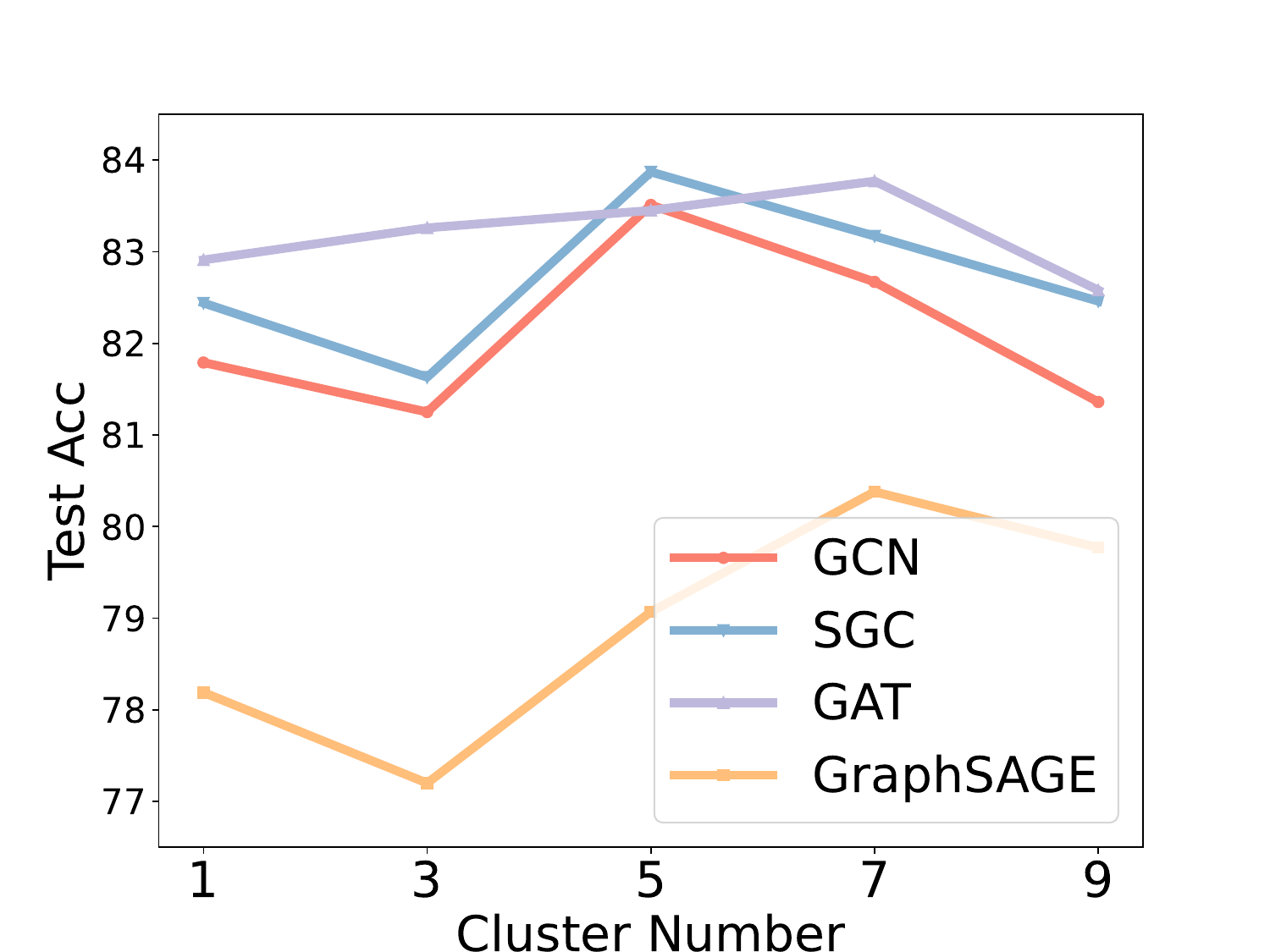}
}\hspace{-15pt}
\subfigure { 
\includegraphics[width=0.65\columnwidth]{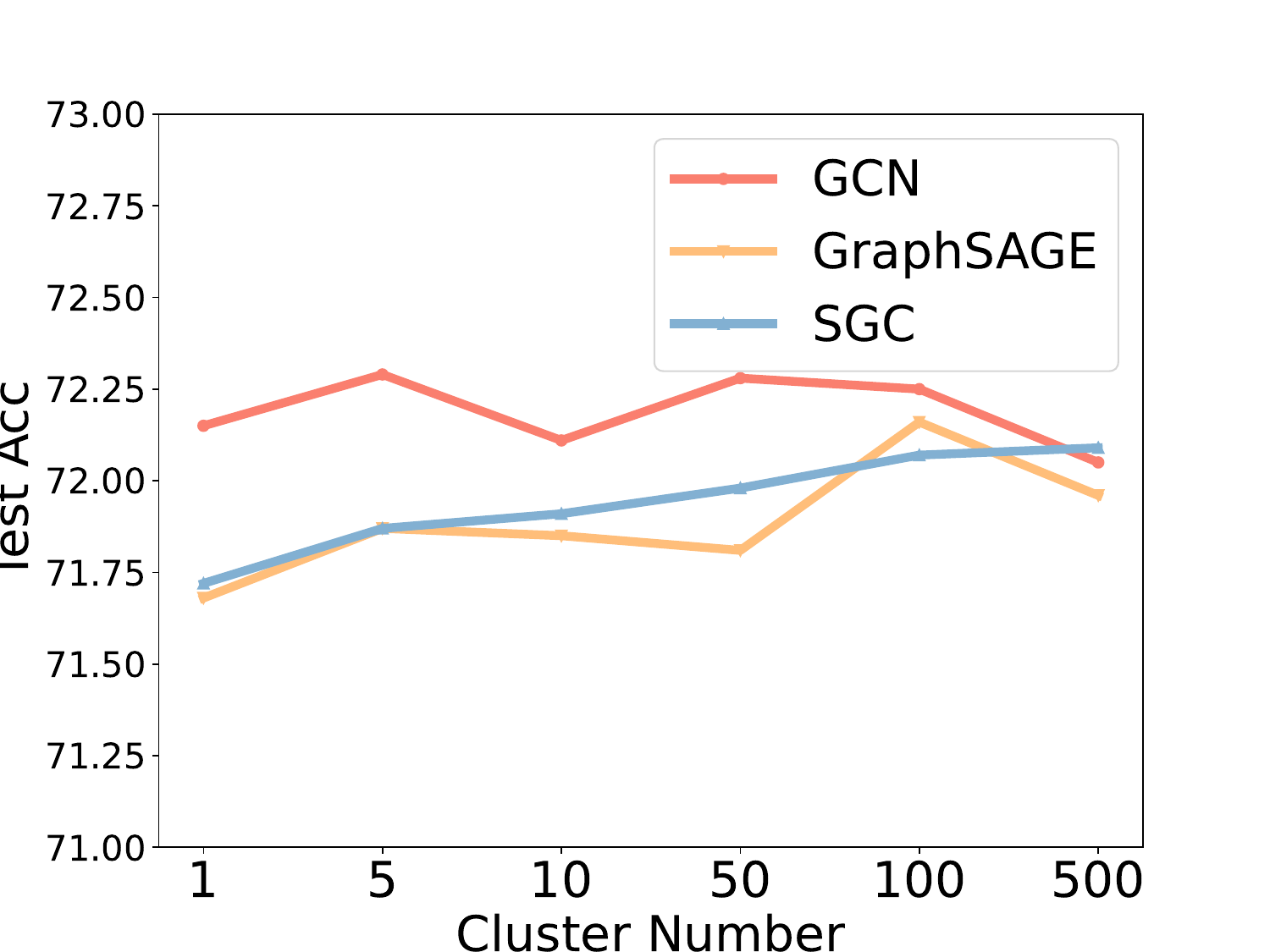}
}\hspace{-15pt}
\subfigure { 
\includegraphics[width=0.65\columnwidth]{cluster_num_arxiv.pdf}
}
\vspace{-12pt}
\caption{Hyperparameter effect of the cluster number in the joint-cluster supervised learning. Note that a/b in Yelp, a denotes cluster number in Cluster\_GCN and GraphSAGE, and b represents cluster number in SIGN, which uses a larger batch size.} 
\label{cluster_num}
\vspace{-10pt}
\end{figure*}



\textbf{Q: Whether the joint learning avoids the overfitting on training set.} The model's generalization ability is commonly measured by the gap between training loss and test loss. The smaller the gap is, the better the model can be free from the overfitting but generalizes on the testing set. We plot such a loss gap in the right part of Figure~\ref{t-sne}, where joint-cluster loss generally has a smaller gap. Since joint-cluster loss avoids the over-confident prediction by referring to cluster information.


\textbf{Q: How is the sensitivity of the joint-cluster learning framework to the cluster number?}
Fig.\ref{cluster_num} shows the hyperparameter effect of the cluster number on both small and large datasets. We observe the joint-cluster loss benefits from a suitable number in a smaller dataset Cora. Yet, we notice the performances are stable as the cluster number changes in larger datasets, such as Arxiv and Yelp. 

\textbf{Q: Does our joint-cluster learning framework require expensive memory cost compared to standard supervised learning framework?}
We examine this question in Table~\ref{table_memory}. It is found that our framework requires little cost on most models except GAT, which brings the non-negligible improvements in node classification accuracy and robustness over adversarial attack. Although GAT requires a higher cost due to its complex attention mechanism, this is still acceptable compared with the benefits. 

\begin{table}[t]
\vspace{-10pt}
\caption{Occupied memory (ratio) of JC loss compared with vanilla cross-entropy loss.}
\label{table_memory}
\vskip 0.05in
\begin{center}
\begin{tabular}{lccr}
\toprule
    Model  & Cora & CiteSeer & PubMed \\
    \midrule
    GCN  & $1.01\times$ &$1.04\times$ &$1.00\times$ \\ 
    SGC  & $1.06\times$ &$1.05\times$ &$1.00\times$ \\
    MLP  & $1.01\times$ &$1.05\times$ &$1.02\times$ \\ 
    SAGE & $1.03\times$ &$1.08\times$ &$1.07\times$ \\ 
    GAT  & $1.70\times$ &$1.39\times$ &$1.56\times$ \\ 
\bottomrule
\end{tabular}
\end{center}
\vspace{-20pt}
\end{table}

\vspace{-5pt}
\section{Conclusion} 
\vspace{-5pt}
In this paper, we hypothesize that the independent conditional distribution of node labels is not in line with the graph-structured data, where nodes tend to connect with ``similar'' neighbors and linked nodes have complicated relationships. Based on the i.i.d assumption, the supervised learning with standard cross-entropy loss fails to fully activate the model's ability in generalizing over a test set as well as defending adversarial attacks. 
Motivated by the label dependencies between nodes and their corresponding clusters, we have presented the joint-cluster supervised learning framework for the training and inference in graph data. This new paradigm learns the joint distribution of nodes and their cluster labels conditioned on their features, and introduces the joint-cluster cross-entropy loss. 
The extensive experiments demonstrate that our model can boost the node classification performance of GNN models and simple MLP architecture compared to the standard supervised learning on a wide range of real-world datasets. The limitations and interesting future work are discussed in Appendix~\ref{appendix_limit}. 

\section*{Acknowledgements}

This work was supported by a grant from the National Science and Technology Major Project (No.2021ZD0112500), and the National Natural Science Foundation of China under grants (No.62372211, No.62272191), and the International Science and Technology Cooperation Program of Jilin Province (No.20230402076GH, No.20240402067GH), and the Science and Technology Development Program of Jilin Province (No.20220201153GX).

\section*{Impact Statement}

We introduce a new supervised learning paradigm for graph-structured data. We focus on extending the node independence assumption to the cluster independence assumption to ensure that the structural information of the graph-structured data is used during loss optimization process.

\nocite{langley00}

\bibliography{example_paper}
\bibliographystyle{icml2024}

\newpage
\appendix
\onecolumn

\section{Algorithm}\label{appendix_algori}

The detailed description of our proposed joint-cluster learning framework.

\begin{algorithm}[h]
   \caption{Joint-Cluster Learning Framework}
   \label{alg}
\begin{algorithmic}
   \STATE {\bfseries Input:} Adjacent matrix $\mathbf{A}$, features matrix $\mathbf{X}$, the set of labeled nodes $\mathbf{V}_{L}$ and their labels $\mathbf{Y}_{L}$, encoder $f_\theta$, classifier $g_\phi$
   \STATE {\bfseries Output:} Predicted labels of unlabeled nodes
   \STATE Partition graph nodes into $M$ \text{clusters} $\mathbf{C}_{1}, \mathbf{C}_{2}, ..., \mathbf{C}_{M}$ \text{by METIS};
   \FOR{$i=1$ {\bfseries to} $M$}
   \STATE $\bar{\mathbf{Y}}_{m} = 1/L_{m}\sum_{k=1}^{L_{m}}\mathbf{Y}_{k};$ \\
     \tikz[baseline]{\node[anchor=base,fill=none,draw=none,text opacity=0.5] {$// \ \text{Calculate cluster\_label according to}\ \mathbf{Y}_{L}.$};}
   \ENDFOR
   \FOR{$i=1$ {\bfseries to} \text{max\_iteration\_epoch}}
   \STATE $\mathbf{Z} = f_\theta\left(\mathbf{A},\mathbf{X}\right);$
   \STATE \tikz[baseline]{\node[anchor=base,fill=none,draw=none,text opacity=0.5] {$// \ \text{Update node embedding.}$};}
    \STATE $\bar{\mathbf{Z}}_{m} = 1/L_{m}\sum_{k=1}^{L_{m}}\mathbf{Z}_{k};$ \\
     \STATE \tikz[baseline]{\node[anchor=base,fill=none,draw=none,text opacity=0.5] {$// \ \text{Update cluster embedding.}$};}\\
     \STATE $\text{Update joint\_embeddings}\ \mathbf{Z}_{jc}\ \text{according to}\ \mathbf{Z}_{L}\ \text{and}\ \bar{\mathbf{Z}}_{m}$\\
     \STATE $\hat{\mathbf{Y}}_{jc} = g_\phi\left(\mathbf{Z}_{jc}\right);$\\
     \STATE \tikz[baseline]{\node[anchor=base,fill=none,draw=none,text opacity=0.5] {$// \ \text{Obtain joint prediction of Node and Cluster.}$};}\\
     \STATE $\mathbf{Y}_{jc} = \mathbf{Y}_{i}\bar{\mathbf{Y}}_{m}^{\top};$\\
     \STATE \tikz[baseline]{\node[anchor=base,fill=none,draw=none,text opacity=0.5] {$// \ \text{Joint label of node and cluster.}$};}\\
     \STATE $\text{Calculate joint-cluster loss}\ \mathbf{L}_{jc} \ \text{according to}\ \mathbf{Y}_{jc}\ \text{and}\ \hat{\mathbf{Y}}_{jc}$\\
     \STATE $\bigtriangledown_{\theta, \phi}\left[\mathcal{L}_{jc} \right]$\\
   \ENDFOR
\end{algorithmic}
\end{algorithm}

\section{The Statistics of Datasets}\label{appendix_data}
Table~\ref{dataset} contains the statistics for the nine datasets used in our experiments for node classification. Experiments are under the single-class and multi-class setting. For single-class classification task, we conduct the experiments on an online social network (LastFMAsia~\citep{LastFM}), a webpage dataset(Wisconsin\footnote{http://www.cs.cmu.edu/afs/cs.cmu.edu/project/theo-11/www/wwkb}), three page-page networks (Facebook~\citep{Facebook}, Chameleon and Squirrel~\citep{cham}) and citation networks, including Cora, CiteSeer, PubMed~\citep{GCN}, DBLP~\citep{DBLP} and ogbn-arxiv~\citep{ogb}. For multi-class classification task, we use businesses types network based on customer reviewers and friendship (Yelp~\citep{graphsaint}), and product network based on buyer reviewers and interactions (Amazon~\citep{graphsaint}). Furthermore, the statistics of the datasets used in adversarial attack in Table~\ref{dataset_attack}.

Next, we will introduce in detail the data split. We follow the standard split proposed by ~\citep{GCN} on three citation networks, including Cora, CiteSeer, and PubMed. For DBLP and Facebook, we use 20 labeled nodes per class as the training set, 30 nodes per class for validation, and the rest for testing. In addition, we conduct the experiments on LastFMAsia and ogbn-arxiv to further evaluate the performance of our proposed joint-cluster loss on imbalanced datasets. For LastFMAsia, we randomly split 25\%/25\%/50\% of nodes for training, validation, and testing. For ogbn-arxiv, we follow the standard split proposed by~\cite{ogb}. For heterophilic graph datasets(Chameleon, Squirrel and Wisconsin), we fellow the data split of~\cite{geom-gcn, yangliang1}. For two large multi-class datasets proposed by~\cite{graphsaint}, including Yelp and Amazon, whose node numbers are 716K and 1598K. Following~\cite{graphsaint}, we use the same data split to stay our focus on the design of the objective function and conduct a fair comparison with independent cross-entropy loss. For robustness experiments, following previous works~\citep{prognn}, we only consider the largest connected component (LCC) in the adversarial graphs, and randomly split 10\%/10\%/80\% of nodes for training, validation, and testing.

\begin{table}
  \caption{Statistics of datasets used in experiments (``m'' stands for multi-class classification, and ``s'' for single-class).}
  \vskip 0.05in
  \label{dataset}
  \centering
  \begin{tabular}{crrrc}
    \toprule
    Datasets & Nodes & Edges & Features & Classes\\
    \midrule
    Cora       & 2,708   & 5,429     & 1,433 & 7(s)\\
    CiteSeer   & 3,327   & 4,732     & 3,703 & 6(s)\\
    PubMed     & 19,717  & 44,338    & 500   & 3(s)\\
    DBLP       & 17,716  & 105,734   & 1,639 & 4(s)\\
    Facebook   & 22,470  & 342,004   & 128   & 4(s)\\
    LastFMAsia & 7,624   & 55,612    & 128   & 18(s)\\
    ogbn-arxiv & 169,343 & 1,166,243 & 128   & 40(s)\\
    Chameleon  & 2,277   & 36,101    & 2,325 & 5(s)\\
    Squirrel   & 5,201   & 217,073   & 2,089 & 5(s)\\
    Wisconsin  & 251     & 499       & 1,703 & 5(s)\\
    Yelp       & 716,847 & 6,977,410 & 300   & 100(m)\\
    Amazon     & 1,598,960 & 132,169,734 & 200 & 107(m)\\
  \bottomrule
\end{tabular}
\end{table}

\begin{table}
  \caption{Following~\cite{prognn}, we only consider the largest connected component (LCC).}
  \vskip 0.05in
  \label{dataset_attack}
  \centering
  \begin{tabular}{crrrc}
    \toprule
    Datasets & Nodes & Edges & Features & Classes\\
    \midrule
    Cora      & 2,485   & 5,069     & 1,433 & 7\\
    CiteSeer  & 2,110   & 3,668     & 3,703 & 6\\
    PubMed    & 19,717  & 44,338    & 500   & 3\\
    Polblogs  & 1,222   & 16,714    & /     & 2\\
  \bottomrule
\end{tabular}
\end{table}

\section{Other Related Work}\label{appendix_related}
\paragraph{Graph neural networks.}
Existing GNNs follow the neighborhood aggregation strategy, which iteratively updates the node representation by aggregating the representations of neighboring nodes and combining them with its representations~\citep{gin,why_pro}. Numerous variants of GNNs have been proposed to achieve outstanding performances in many graph-based tasks, such as graph clustering~\citep{cluster1, cluster2}, node classification~\citep{GCN,Adagcl,gcl_eu} and graph classification~\citep{graphc2, graphc1}. To deal with large-scale graph, researchers have proposed some scalable graph learning methods~\citep{cluster_gcn, rethinklarge}. 

\paragraph{Graph adversarial attack.}
Graph adversarial attack refers to the process of manipulating or perturbing the nodes, edges, or features in a graph to deceive or mislead graph-based learning models\citep{chen2020survey,jin2020adversarial}. 
These attacks can be categorized into different types, such as structural attacks that modify the graph topology~\citep{xu2019topology,wang2019attacking,li2020adversarial}, feature-based attacks that manipulate node features~\citep{liu2022adversarial}, and hybrid attacks that combine both~\citep{zhang2022robust,xie2022revisiting,ma2022adversarial}. 
Compared with cross-entropy loss, our joint-cluster loss can refer to similar nodes in the process of loss optimization and inference, which can effectively alleviate the impact of graph attacks.


\section{Description of Backbone Models}\label{appendix_backbone}
We evaluate our joint-cluster supervised learning framework on differnet GNN models and scalable graph learning backbones:
\begin{itemize}[wide=0pt, leftmargin=\dimexpr\labelwidth + 2\labelsep\relax]
    \item \textbf{GCN~\citep{GCN}:} GCN is a convolutional neural network which utilizes the structural information of graphs by message passing mechanism.
    \item  \textbf{SGC~\citep{SGC}:} SGC eliminates the nonlinearities of GCN and collapses the weight matrix into a weight matrix.
    \item \textbf{MLP:} MLP is a simple neural network that maps a set of input vectors to a set of output vectors.
    \item \textbf{GAT~\citep{GAT}:} GAT learns edge weights in graph domain through the attention mechanism and achieves significant performance.
    \item \textbf{GraphSAGE~\citep{GraphSage}:} Graphsage obtains neighbor nodes through sampling strategies and expresses node representation through neighbor aggregation operations.
    \item \textbf{Cluster\_GCN~\citep{cluster_gcn}:} Cluster\_GCN is a fast and efficient mini-batch training algorithm that preserve structural information within a batch by exploiting the graph clustering structure.
    \item \textbf{SIGN~\citep{sign}:} SIGN is amenable to efficient precomputation by using graph convolutional filters of different size, achieving fast training and inference.
\end{itemize}

\section{Detailed Explanations}\label{more_explanations}
\paragraph{How to quantify and define the ``joint distribution'' of nodes and clusters?} 
According to Eq. (4a)-(4e) at Page $3$, the conditional joint distribution is defined as $p(y,\bar{y} \mid z, \bar{z})$, where $y$ and $\bar{y}$ are label variables of node and cluster, respectively; $z$ and $\bar{z}$ are representation vector variables of them. To facilitate the empirical evaluation of joint distribution, we cluster the underlying graph into many groups and only account the joint probability of nodes and their corresponding cluster, since the random walks startign from a node (i.e., node's flow influence) are often trapped in the local cluster~\citep{cluster_help}. Thus the empirical estimation of joint distribution on input graph is $\prod_{m=1}^{M} \prod_{i=1}^{|C_m|} p(y_{i}, \bar{y}\_{m} \mid z\_{i}, \bar{z}\_{m}; \theta)$, where $M$ is the number of clusters, $|\mathcal{C}_m|$ is the number of nodes within the $m$-th cluster. To represent the label and representation variables of cluster, we use the means of training nodes within the cluster, i.e., $\bar{y}_{m}$ and $\bar{z}_{m}$, for the computation purpose. As illustrated in Fig.~\ref{F3}, we use a fully-connected networks with paramethers $\theta$ to approximate the joint distribution, which takes the concatenate input of node and cluster representations ($z_{i}$ and $\bar{z}_{m}$) and output the joint label with size of $\mathbb{R}^{c\times c}$, where $c$ is the number of class categories.

\paragraph{How this new assumption concretely impacts the learning process?}
Motivated from natural instincts that advanced intelligence makes decision with similar experienced samples as reference, we propose the joint distribution to calibrate the node class decision via considering its multi-hop neighborhood distribution. Concretely, the joint probability of node $i$ and cluster $m$ is defined as $p_{i,m} = y_i\bar{y}_{m}^T \in \mathbb{R}^{c\times c}$, and the sum of elements in $p_{i,m}$ equal to one. Taking the binary classification as example, supposing that ground-truth of two training nodes are $y_1 = [0, 1]$ and $y_2 = [0, 1]$ at cluster with average label $\bar{y}_m = [0.2, 0.8]$, we have the  joint probabilities of $[[0, 0], [0.2, 0.8]]$ and $[[0.2, 0.8], [0, 0]]$. According to cross-entropy loss at Eq. (5), with the calibration of cluster label, the differentiable encoder will be regualized to assign prediction confidences on other reference classes (i.e., the non-diagonal entries at $c\times c$ prob estimation $\hat{p}_{i,m}$). In other word, the learning process is optimized to learn a smooth estimation of $\hat{p}_{i,m}$ and avoids over-fitting, where each element at $\hat{p}_{i,m}$ is supervised to encode the joint probability of specific class tuple and has value highly smaller than one.  At the inference phase, the prob estimation $\hat{p}_{i}$ is obtained by marginalizing along the cluster dimension at $\hat{p}_{i, m}$, which makes use of all the cluster-based joint dependencies to infer a cautious class decision. In the experiment, we have observed the smooth estimation of joint probability and the marginalizing-based inference signigicantly enhance classification accuracy, imbalanced classification, and robustness, via tackling the over-fitting in learning process.

\paragraph{Whether there is an explicit or potential connection between labels and clusters?}
The connection beween labels and clusters depend on the graph type: homogeneous or heterogeneuos garph. While node labels in the same cluster tend to be same at homogeneous graph, they show distinctly at heterogeneous graph. But our framework does not assume on either of the connections and prefer the division result where the node labels within a cluster are not completely consistent with each other. That is because one can leverage the cluster's diverse labels to calibrate each node as explained before. Our experimental results show that our joint-cluster framework achieve promising accuracy in both homogeneous or heterogeneuos garphs, without relying the connection assumption.

\section{Multi-Class Task Design} 
We introduced the framework design of single-class classification task in the paper. In short, for the single-class setting, joint-cluster learning framework expands a $c$-class classification task into a $c^{2}$-class classification task. The multi-class setting is slightly different from single-class. The number of clsses $c$ in the multi-classification task represents $c$ binary classification tasks. We extend each two-class classification task to a four-class classification task for nodes and clusters, and use cross-entropy loss to optimize each four-class classification. So the output dimension of the classifier is $4c$.

\section{Implementation}
Following the experimental settings of original papers, for GAT\footnote{https://github.com/pyg-team/pytorch\_geometric/blob/master/examples/gat.py}, we choose the model parameters by utilizing an early stopping strategy with a patience of 100 epochs on classification loss. For other GNN models\footnote{https://github.com/tkipf/pygcn}\footnote{https://github.com/Tiiiger/SGC}, we utilize the model parameters which perform best on the validation set for testing. The remaining hyper-parameters including learning rate, dropout and weight decay are tuned for different models. Scalable graph learning methods are executed based on the official examples of PyTorch Geometric\footnote{https://github.com/pyg-team/pytorch\_geometric/blob/master/examples/cluster\_gcn\_ppi.py}\footnote{https://github.com/pyg-team/pytorch\_geometric/blob/master/examples/sign.py}. We further implement joint-cluster loss over each backbone framework. Because Graphsage and SIGN divide the batch, it is impossible to guarantee that the nodes in the same batch are adjacent. Therefore, in order to ensure fairness, for the joint-cluster loss of the large-scale graph learning methods, we use the manner of randomly assigning clusters to the nodes.

\begin{table*}[t]
  \caption{Test Accuracy (\%) for different models on five datasets. In addition, we show the best results in bold. We run 10 times and report the mean ± standard deviation. CE denotes the standard cross-entropy loss, IC denotes in-context learning strategy, and JC denotes our joint-cluster learning framework.}
  \vskip 0.05in
  \label{ic_res}
  \centering
  \setlength{\tabcolsep}{9.5pt}
  \begin{tabular}{cccccccc}
    \toprule
    Model  & Loss  & Cora & CiteSeer & PubMed  & DBLP  & Facebook\\
    \midrule
    \multirow{3}{*}{GCN} & CE     &$81.70_{\pm0.65}$  &$71.43_{\pm0.47}$    &$79.06_{\pm0.32}$   &$74.30_{\pm 1.94}$  &$73.91_{\pm 1.40}$\\
                         & IC     &$81.56_{\pm0.25}$  &$70.08_{\pm0.56}$    &$79.37_{\pm0.46}$   &$72.53_{\pm2.55}$  &$70.35_{\pm1.86}$\\
                         & JC    &$\mathbf{83.51_{\pm 0.35}}$  &$\mathbf{72.97_{\pm 0.55}}$    &$\mathbf{79.80_{\pm 0.19}}$  &$\mathbf{75.10_{ \pm 1.63}}$  &$\mathbf{74.64_{\pm 1.75}}$\\
    \hline
     \multirow{3}{*}{SGC} & CE   &$81.68_{\pm0.52}$  &$71.85_{\pm0.39}$    &$78.70_{\pm0.38}$  & $74.30_{\pm 2.12}$ &$ 74.13_{\pm 2.13}$\\
                         & IC     &$81.87_{\pm0.51}$  &$69.41_{\pm0.79}$    &$79.20_{\pm0.43}$   &$71.40_{\pm 1.07}$  &$68.15_{\pm 3.30}$\\
                          & JC  &$\mathbf{83.87_{\pm 0.79}}$  &$\mathbf{72.92_{\pm 0.16}}$    &$\mathbf{79.97_{\pm 0.25}}$   & $\mathbf{74.87_{\pm 1.81}}$ & $\mathbf{74.74_{\pm 1.96}}$   \\
    \hline
    \multirow{2}{*}{SAGE} & CE     &$79.96 _{\pm 0.44}$  &$69.94_{ \pm 0.93}$    &$78.37_{ \pm 0.72}$   &$70.59_{\pm 1.46}$  &$70.95_{\pm 2.26}$\\ 
                         & IC     &$78.70 _{\pm 1.13}$  &$67.52_{ \pm 0.96}$    &$78.50_{ \pm 0.58}$   &$70.17_{\pm 3.29}$  &$69.75_{\pm 1.75}$\\ 
                         & JC    &$\mathbf{80.81_{\pm 0.63}}$  &$\mathbf{70.54_{\pm 1.49}}$    &$\mathbf{79.50_{\pm 1.02}}$  &$\mathbf{71.87_{ \pm 2.07}}$  &$\mathbf{71.59 _{\pm 1.78}}$\\

    \hline
    \multirow{2}{*}{GAT} & CE    & $83.22_{\pm 0.29}$ & $71.06_{ \pm 0.40}$ & $78.54_{ \pm 0.63}$   & $75.32_{\pm 2.62}$   & $76.34_{\pm 2.26}$\\
          & IC     &$83.21_{\pm0.32}$  &$71.43_{\pm0.47}$    &$78.38_{\pm0.22}$   
          &$74.10_{\pm 1.59}$
          &$72.49_{\pm 2.34}$\\
                         & JC   & $\mathbf{83.77_{ \pm 0.44}}$ & $\mathbf{70.18 _{\pm 0.86}}$ & $\mathbf{79.35_{ \pm 0.47}}$  & $\mathbf{76.92_{\pm 1.59}}$  & $\mathbf{77.46_{\pm 2.30}}$\\
    \hline
    \multirow{2}{*}{MLP} & CE   & $58.65_{ \pm 0.97}$ & $60.41_{ \pm 0.56}$ & $73.27_{ \pm 0.35}$ & $47.95_{ \pm 3.97}$  & $55.34_{ \pm 2.60}$\\
        & IC     &$64.27_{\pm0.43}$  &$62.27_{\pm1.69}$    &$75.74_{\pm0.46}$   &$58.83_{\pm 2.31}$  &$56.53_{\pm 3.20}$\\
                        & JC   & $\mathbf{67.19_{ \pm 0.62}}$ & $\mathbf{63.23_{ \pm 0.87}}$ & $\mathbf{75.92_{\pm 0.39}}$  & $\mathbf{61.16_{ \pm 3.63}}$  & $\mathbf{56.62_{\pm 2.42}}$\\
  \bottomrule
\end{tabular}
\end{table*}

\section{Additional Experiments}

\paragraph{In-context learning.}\label{appendix_ic}
We conduct experiments to demonstrate the effect of joint distribution modeling in joint-cluster supervised learning framework. For each experiment, we compare the model trained by  standard supervised learning, in-context strategy and joint-cluster learning framework. For in-context learning, we use the same input as the joint-cluster framework, the output is a c-dimensional vector, and the node label is used as the ground truth. As shown in Table~\ref{ic_res}, we observe that in-context strategy does not get a stable accuracy improvement. We guess that in-context strategy requires the cluster label should be sharp and the node label should be consistent with the cluster label, which will cause the model to be limited by the division of clusters. Our joint-cluster framework learns the joint distribution of nodes and clusters, which will learn potentially complex relationships between nodes, not just similarities.

\paragraph{Cluster mixup.}
It is worth noting that employing the clusters obtained by mixing training samples for training through independent cross-entropy loss can potentially bring improvements, which is similar to mixup. We need to provide further comparison to clarify the advantages of JC loss. 
Therefore, we directly apply the independent cross entropy loss after obtaining the features $\bar{z}_{m}$ and labels $\bar{y}_{m}$ of the cluster to obtain the cluster training loss $L_{clu}$, and the overall loss function can be presented as $\mathcal{L} = \mathcal{L}_{ce} + \beta \cdot \mathcal{L}_{ce}$, where $\beta$ > 0 is tuning parameter to weight the importance of the cluster training loss. As shown in the table~\ref{cluster_mixup}, experimental results indicate that applying independent cross-entropy loss on mixup cluster samples yields some improvements, primarily attributed to the data augmentation of mixup samples and the challenge of fitting mixed samples to bring generalization capability. However, the JC loss demonstrates more pronounced enhancements. Different from the independent training on mixup cluster samples, our joint-cluster supervised learning provides a paradigm to end-to-end learn the joint distribution of target node and its cluster, which brings generalization benefit to infer every node classes via using cluster distribution as reference signals.

\begin{table}[t]
  \caption{Test accuracy (\%) on  citation networks, CE denotes the standard cross-entropy loss, Mixup denotes cluster mixup training strategy, and JC denotes our joint-cluster learning framework.}
  \vskip 0.05in
  \label{cluster_mixup}
  \centering
  \begin{tabular}{cccc}
    \toprule
    Methods   & Cora & CiteSeer & PubMed \\
    \midrule
    GCN+CE    & $81.70_{\pm 0.65}$ & $71.43_{\pm 0.47}$ & $79.06_{\pm 0.32}$ \\
    GCN+Mixup & $82.43_{\pm 0.44}$ & $71.68_{\pm 0.51}$ & $79.36_{\pm 0.17}$ \\
    GCN+JC    & $\mathbf{83.51_{\pm 0.35}}$ & $\mathbf{72.97_{\pm 0.55}}$ & $\mathbf{79.80_{\pm 0.19}}$ \\
  \bottomrule
\end{tabular}
\end{table}

\begin{table*}[t]
  \centering
  \setlength{\tabcolsep}{4pt}
  \caption{The efficiency analysis of the training time and training memory.} 
  \vskip 0.05in
  \label{effiency}
  \begin{tabular}{cccccccc}
   \toprule
   \multirow{2}{*}{Datasets} & \multirow{2}{*}{Loss} & \multicolumn{2}{c}{Cora} & \multicolumn{2}{c}{CiteSeer} & \multicolumn{2}{c}{PubMed} \\
   \cmidrule{3-8} &    & Time(s)  & Memory(MB)  & Time(s)   & Memory(MB)  & Time(s)   & Memory(MB)  \\
   \hline
   \multirow{2}{*}{GCN} & CE     &$0.002$  &$88.29$   &$0.002$  &$184.90$   &$0.016$  &$3061.18$ \\
                         & JC    &$0.004$  &$89.16$   &$0.005$  &$191.53$   &$0.018$  &$3065.76$ \\
    \hline
   \multirow{2}{*}{SGC} & CE     &$0.002$  &$60.55$   &$0.002$  &$142.67$   &$0.002$  &$1577.18$ \\
                         & JC    &$0.003$  &$64.39$   &$0.005$  &$149.24$   &$0.007$  &$1581.75$\\
    \hline
   \multirow{2}{*}{MLP} & CE     &$0.001$  &$60.55$   &$0.002$  &$142.67$   &$0.002$  &$1577.18$ \\
                         & JC    &$0.004$  &$61.15$   &$0.004$  &$149.24$   &$0.006$  &$1603.10$\\
    \hline
   \multirow{2}{*}{SAGE} & CE     &$0.005$  &$49.54$   &$0.005$  &$148.00$   &$0.006$  &$147.51$ \\
                         & JC    &$0.008$  &$51.16$   &$0.008$  &$159.22$   &$0.008$  &$157.84$\\
    \hline
   \multirow{2}{*}{GAT} & CE     &$0.005$  &$61.60$   &$0.005$  &$157.43$   &$0.006$  &$246.93$ \\
                         & JC    &$0.007$  &$104.99$   &$0.008$  &$218.44$   &$0.011$  &$385.76$\\
   \bottomrule
  \end{tabular}%
 \end{table*}

\paragraph{Efficiency analysis.}
We use METIS to efficiently perform cluster division at pre-processing stage for small graphs with thousands of nodes, which takes less than five seconds. For the batch training on large graphs, we use random clustering on sampled training nodes and do not require clustering time cost. Next we show the training time and training memory per epoch for vanilla cross-entropy (CE) loss and our joint-cluster (JC) loss in Table~\ref{effiency}. It is found that computational time overhead and memory cost are extremely marginal, which brings the non-negligible improvements in node classification accuracy and robustness over adversarial attack.

\begin{figure}[h]
\centering 
\vspace{-15pt}
\subfigure[Cora] { 
\includegraphics[width=0.3\columnwidth]{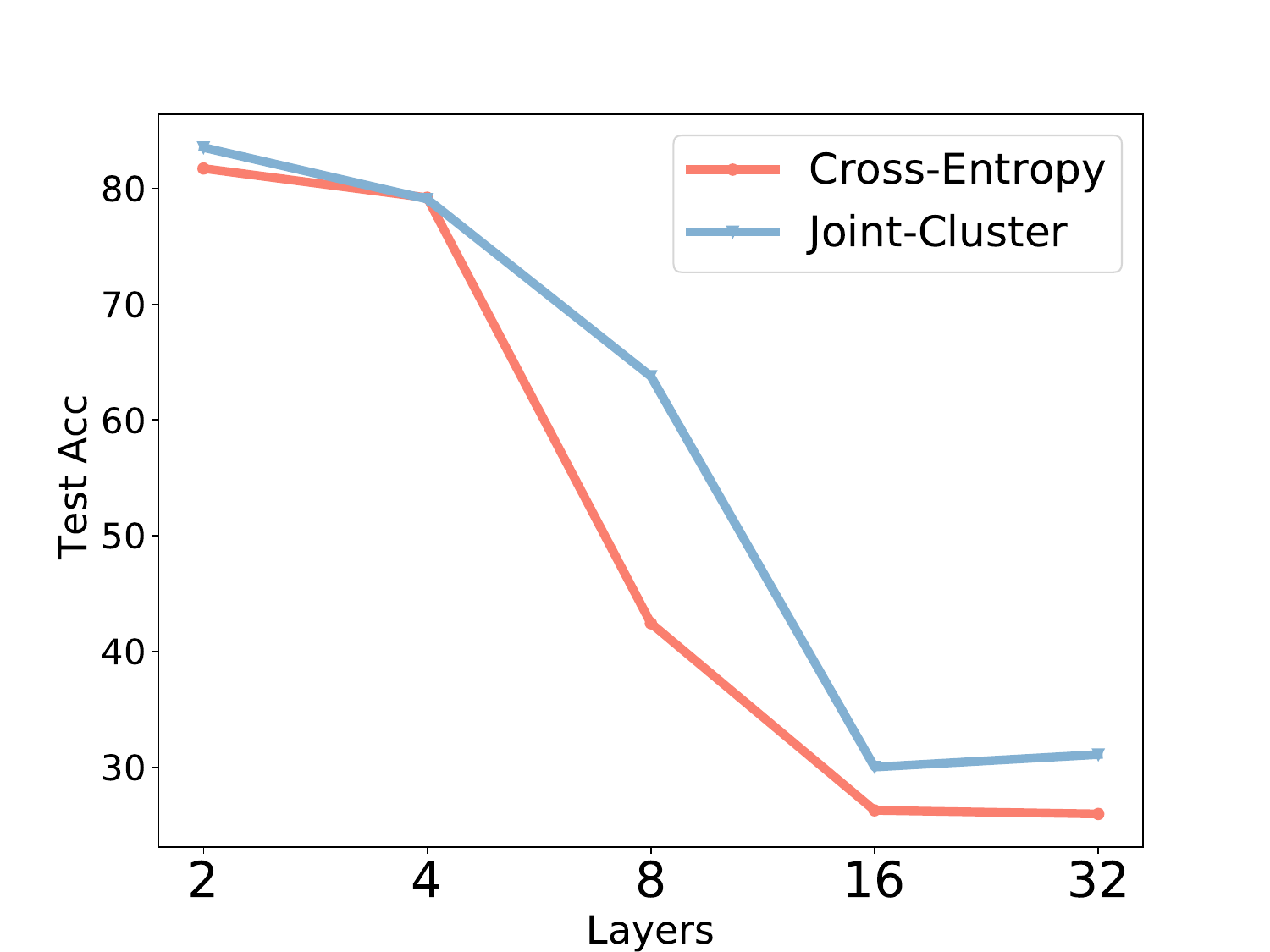}
}\hspace{-7mm}
\subfigure[CiteSeer] { 
\includegraphics[width=0.3\columnwidth]{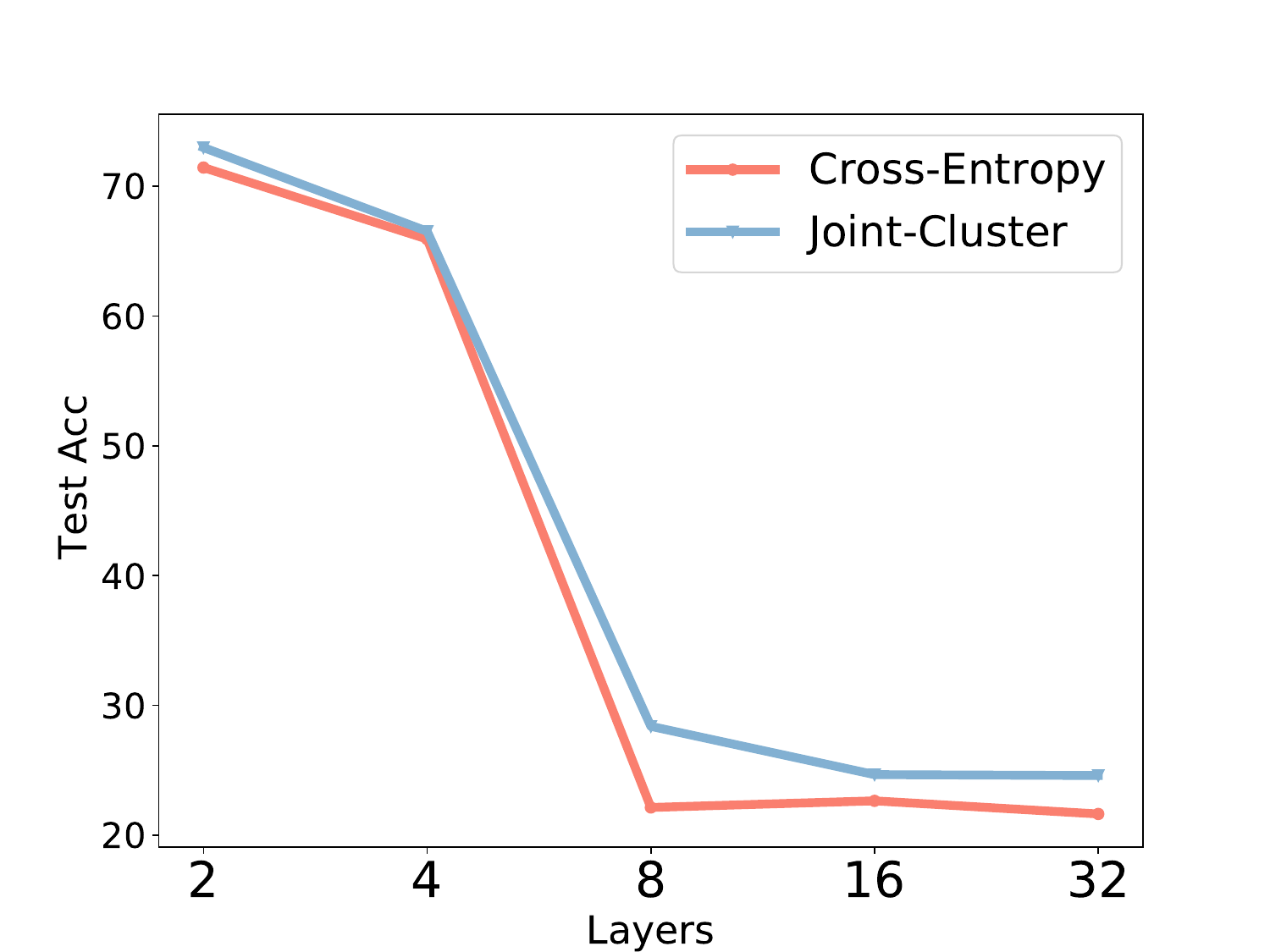}
}\hspace{-7mm}
\subfigure[PubMed] { 
\includegraphics[width=0.3\columnwidth]{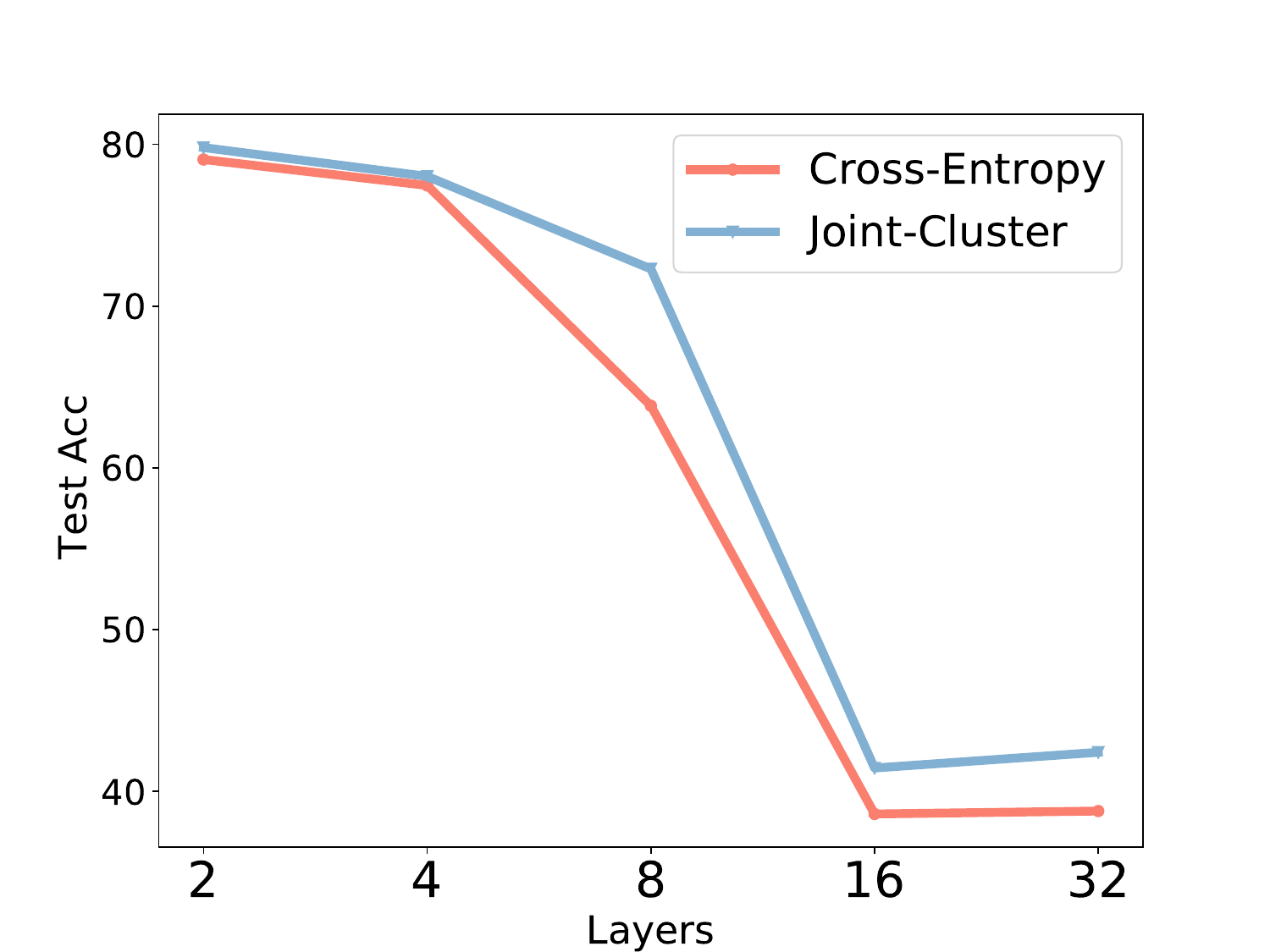}
}
\caption{Over-smoothing analysis about the model depth for node classification.}
\label{over-smooth}
\vspace{-0.6cm}
\end{figure}

\paragraph{Over-smoothing.}
Over-smoothing which suggests that as the number of layers increases, the representations of the nodes in GCN are inclined to converge to a certain value and thus become indistinguishable~\citep{oversmooth1}. A number of models have been recently proposed to alleviate the over-smoothing issue, including skip connection~\citep{GCNII}, orthogonal regularization~\citep{oversmooth3, oversmooth2, yangliang2} and edge adjustment~\citep{oversmooth4}. Our framework can alleviate over-smoothing. As shown in Figure 2 of manuscript, in a 8-layer GCN, our framework can exhibit 2D projection of node embeddings with more coherent shapes of clusters. In addition, we leverage GCN as the backbone networks, and compare joint-cluster loss with cross-entropy loss by considering the layer numbers of 2, 4, 8, 16, and 32. As shown in the Figure~\ref{over-smooth}, our approach almost delivers the better node classification accuracies. That is because our framework separates the node distribution modeling of different clusters, which could relieve the over-smoothing issue to some extent.

 \begin{table*}[t]
  \centering
  \setlength{\tabcolsep}{4pt}
  \caption{Classification accuracy(\%) among different models with varying numbers of labels on three citation networks.} 
  \vskip 0.05in
  \label{few-label}
  \begin{tabular}{cccccccc}
   \toprule
   \multirow{2}{*}{Datasets} & \multirow{2}{*}{Label} & \multicolumn{2}{c}{GCN} & \multicolumn{2}{c}{SGC} & \multicolumn{2}{c}{MLP} \\
   \cmidrule{3-8} &    & CE  & JC  & CE   & JC  & CE   & JC  \\
   \hline
   \multirow{2}{*}{Cora} & 5     &$71.80_{\pm2.48}$  &$\mathbf{72.38_{\pm2.73}}$   &$72.61_{\pm2.59}$  &$\mathbf{73.67_{\pm2.72}}$   &$44.37_{\pm3.55}$  &$\mathbf{51.25_{\pm2.76}}$ \\
                         & 10    &$78.08_{\pm1.13}$  &$\mathbf{78.86_{\pm1.45}}$   &$79.34_{\pm0.70}$  &$\mathbf{79.49_{\pm1.73}}$   &$53.66_{\pm1.55}$  &$\mathbf{59.19_{\pm3.28}}$ \\
    \hline
   \multirow{2}{*}{CiteSeer} & 5     &$57.91_{\pm6.18}$  &$\mathbf{60.98_{\pm5.08}}$   &$56.67_{\pm5.32}$  &$\mathbf{61.34_{\pm6.56}}$   &$45.52_{\pm3.83}$  &$\mathbf{46.84_{\pm5.00}}$ \\
                         & 10    &$68.22_{\pm1.36}$  &$\mathbf{69.92_{\pm1.28}}$   &$68.62_{\pm1.13}$  &$\mathbf{69.88_{\pm1.14}}$   &$53.47_{\pm2.52}$  &$\mathbf{55.85_{\pm2.22}}$\\
    \hline
   \multirow{2}{*}{PubMed} & 5     &$70.36_{\pm4.79}$  &$\mathbf{72.00_{\pm4.96}}$   &$70.77_{\pm3.77}$  &$\mathbf{71.85_{\pm4.65}}$   &$63.45_{\pm2.27}$  &$\mathbf{67.84_{\pm3.76}}$ \\
                         & 10    &$74.94_{\pm2.61}$  &$\mathbf{75.65_{\pm1.71}}$   &$75.38_{\pm2.83}$  &$\mathbf{75.79_{\pm2.17}}$   &$67.28_{\pm1.41}$  &$\mathbf{71.72_{\pm2.14}}$\\
   \bottomrule
  \end{tabular}%
 \end{table*}

\paragraph{Label scarcity setting.}
To verify the performance of our joint\_cluster loss under label scarcity environment, we conduct experiments on three citation networks, and the number of labeled nodes is very small. For all citation networks, we randomly selected five and ten labeled nodes per class as the training set, leaving the validation and test sets unchanged at 500 nodes for validation and 1000 nodes for testing respectively. As shown in~\ref{few-label}, onsidering each backbone model (i.e., GCN, SGC and MLP), our joint-cluster loss consistently diliver a much higher node classification accuracy.

\paragraph{Clustering methods.}
We compare two clustering methods, K-means and METIS, on citation networks using the GCN encoder, where the number of clusters was set to 5 and the number of classes of corresponding dataset. In table~\ref{cluster_methods_num}, it is found METIS achieves a better accuracy. In table~\ref{clustering-methods}, we list the node classification accuracy, the number of preserved within-cluster links, the number of dropped between-cluster links and the rate of within-cluster links to between-cluster links during clusering. As shown in table~\ref{clustering-methods}, we find that MSTIS achieves better classification accuracy of nodes. The intuitive reason is METIS aims to construct the vertex partitions such that within-cluster links are much denser than between-cluster links to capture the community structure of the graph. As validated by the clustering metrics, the preserved number of within-cluster links in METIS are more than that of K-means, while the dropped number of between-cluster links is smaller. The reduction of between-cluster link breaking can lower the approximation error of joint distribution in Eq.(3b). METIS, which divides clusters based on graph structure, is more suitable for joint-cluster supervised learning framework than K-means, which divides clusters based on features.

 \begin{table*}[t]
  \centering
  \setlength{\tabcolsep}{4pt}
  \caption{Hyperparameter effect of the cluster number in the joint-cluster supervised learning with different clustering methods.} 
  \vskip 0.05in
  \label{cluster_methods_num}
  \begin{tabular}{ccccccc}
   \toprule
   \multirow{2}{*}{Methods} &\multicolumn{2}{c}{Cora} &\multicolumn{2}{c}{CiteSeer} &\multicolumn{2}{c}{PubMed}\\
   \cmidrule{2-7}  & 5  & 7  & 5   & 6  & 5   & 3  \\
   \hline
   K-Means    &$82.35_{\pm 0.72}$  &$82.08_{\pm 0.77}$  &$72.30_{\pm 0.54}$  &$72.07_{\pm 0.81}$  &$79.43_{\pm 0.54}$  &$79.28_{\pm 0.47}$ \\
   METIS      &$83.51_{\pm 0.35}$  &$82.76_{\pm 0.61}$  &$72.87_{\pm 0.24}$  &$72.63_{\pm 0.52}$  &$79.80_{\pm 0.19}$  &$79.54_{\pm 0.21}$ \\
   \bottomrule
  \end{tabular}%
 \end{table*}
 
 \begin{table*}[t]
  \centering
  \setlength{\tabcolsep}{4pt}
  \caption{The relationship between clustering performance and final performance with different clustering methods.} 
  \vskip 0.05in
  \label{clustering-methods}
  \begin{tabular}{cccccc}
   \toprule
     Datasets  & Methods  & Acc & W-C links & B-C links  & Rate \\
   \hline
   \multirow{2}{*}{Cora} & K-Means    &$82.43_{\pm0.78}$  &$2857$   &$2421$  &$1.18$ \\
                         & METIS    &$\mathbf{83.51_{\pm0.35}}$  &$4910$   &$368$  &$13.34$ \\
    \hline
   \multirow{2}{*}{CiteSeer} & K-Means    &$72.18_{\pm0.52}$  &$2740$   &$1812$  &$1.51$ \\
                         & METIS    &$\mathbf{72.79_{\pm055}}$  &$4430$   &$122$  &$36.31$ \\
    \hline
   \multirow{2}{*}{PubMed} & K-Means  &$79.35_{\pm0.51}$  &$35452$   &$8872$  &$4.00$ \\
                         & METIS     &$\mathbf{79.80_{\pm0.19}}$  &$41028$   &$3296$  &$12.45$ \\
   \bottomrule
  \end{tabular}
 \end{table*}

\paragraph{More GNN backbones.}
To demonstrates the effectiveness and generalizability of our joint-cluster learning framework, we further validate in the more backbones, including the more advanced backbone GCNII~\citep{GCNII} and the spectral-based ChebNet~\citep{ChebNet}. We compare with vanilla cross-entropy (CE) loss on citation networks. As shown in table~\ref{more-backbones}, for ChebNet, we can conclude that our joint-cluster distribution learning consistently dilivers the superior performance in ChebNet. The intuitive reason is the node label dependency is an underlying and common phenomenon in graph data. 
For GCNII, our joint-cluster (JC) loss dilivers a much higher node classification accuracy except Cora. That is because the proposed joint-cluster distribution learning leverages the node label dependencies within cluster to make cautious inference and better adapt to graph data. GCNII stacks 64 layers of graph convolutional networks on Cora, which is deeper than the layer numbers used in other datasets. Due to the smaller size of Cora, the message passing to a target node in 64-layer GCNII could originate from  entire graph, where the stacking of joint-cluster inference will make the node labels too over-smooth in the same cluster to damage the classification accuracy. 

 \begin{table*}[t]
  \caption{Classification accuracy(\%) among different models on citation networks.}
  \label{more-backbones}
  \centering
  \vskip 0.05in
  \setlength{\tabcolsep}{9.5pt}
  \begin{tabular}{cccccccc}
    \toprule
    Model  & Loss  & Cora & CiteSeer & PubMed  & Arxiv\\
    \midrule
    \multirow{2}{*}{ChebNet} & CE     &$80.54_{\pm0.66}$  &$70.26_{\pm0.59}$    &$78.63_{\pm0.35}$   & $-$ \\
                         & JC    &$\mathbf{81.29_{\pm0.92}}$  &$\mathbf{70.99_{\pm0.64}}$ &$\mathbf{79.18_{\pm 0.62}}$  & $-$  \\
    \hline
     \multirow{2}{*}{GCNII} & CE   &$\mathbf{85.35_{\pm0.56}}$  &$73.37_{\pm0.65}$    &$80.36_{\pm0.43}$  & $72.74_{\pm0.16}$ \\
                          & JC  &$84.78_{\pm0.76}$  &$\mathbf{73.96_{\pm0.68}}$  &$\mathbf{80.95_{\pm0.33}}$   & $\mathbf{73.04_{\pm0.17}}$  \\
  \bottomrule
\end{tabular}
\end{table*}

\paragraph{Adversarial attacks.}
To fully demonstrate the effectiveness of our approach against adversarial attacks, we compare the CE and JC losses in response to random attacks~\citep{prognn}, which randomly injects fake edges into the graph. We first use the attack method to poison the graph. We then train JC loss and CE loss on the poisoned graph and evaluate the node classification of these two losses on various backbones. We evaluate how our approach behaves under different ratios of random noises from 20\% to 100\% with a step size of 20\%, where the noise ratio represents the ratio of added false edges to real edges. As shown in table~\ref{random_attack}, our joint-cluster loss consistently outperforms cross-entropy loss under all perturbation rates. Compared with the independent cross-entropy loss that only uses neighborhood information, our joint-cluster learning framework refers to a larger range of cluster information, which enables the model to reduce the interference caused by wrong neighbors by referring to more information.

\begin{table*}[t]
  \centering
  \setlength{\tabcolsep}{4pt}
  \caption{Test accuracy (\%) under random attacks, where Ptb Rate means the ratio of added false edges to real edges.} 
  \vskip 0.05in
  \label{random_attack}
  \begin{tabular}{cccccccc}
   \toprule
   \multirow{2}{*}{Datasets} & \multirow{2}{*}{Ptb Rate(\%)} & \multicolumn{2}{c}{GCN} & \multicolumn{2}{c}{SGC} & \multicolumn{2}{c}{GAT} \\
   \cmidrule{3-8} &    & CE  & JC  & CE   & JC  & CE   & JC  \\
   \hline
   \multirow{5}{*}{Cora}
                         & 20\%    &$79.44_{\pm 0.50}$  &$\mathbf{80.03_{\pm 0.31}}$  &$79.84 _{\pm 0.02}$  &$\mathbf{80.36_{ \pm 0.04}}$  &$79.72_{\pm 0.53}$  &$\mathbf{80.32_{ \pm 0.71}}$\\
                         & 40\%    &$76.38_{\pm 0.82}$  &$\mathbf{77.02_{\pm 0.23}}$  &$77.15 _{\pm 0.02}$  &$\mathbf{77.53_{ \pm 0.06}}$  &$76.21_{\pm 0.70}$  &$\mathbf{77.07_{ \pm 0.60}}$\\
                         & 60\%    &$74.85_{\pm 0.63}$  &$\mathbf{75.93_{\pm 0.19}}$  &$74.25 _{\pm 0.05}$  &$\mathbf{75.19_{ \pm 0.04}}$  &$73.42 _{\pm0.61}$  &$\mathbf{74.78_{ \pm 0.95}}$\\
                         & 80\%    &$73.39_{\pm 0.58}$  &$\mathbf{74.64_{\pm 0.58}}$  &$73.07 _{\pm 0.05}$  &$\mathbf{73.21_{ \pm 0.03}}$  &$72.13 _{\pm 1.04}$  &$\mathbf{72.92_{ \pm 0.62}}$\\
                         & 100\%    &$71.47_{\pm 0.71}$  &$\mathbf{73.03_{\pm 0.30}}$  &$70.48 _{\pm 0.05}$  &$\mathbf{71.13_{ \pm 0.03}}$  &$69.96 _{\pm 1.11}$  &$\mathbf{70.31_{ \pm 0.77}}$\\
   \hline
   \multirow{5}{*}{CiteSeer} 
                         & 20\%    &$68.76_{\pm 0.67}$  &$\mathbf{69.51_{\pm 0.79}}$  &$69.43_{\pm 0.06}$  &$\mathbf{70.79_{\pm 0.08}}$  &$69.72_{\pm 1.22}$  &$\mathbf{69.87_{\pm 0.36}}$\\
                         & 40\%    &$65.27_{\pm 0.42}$  &$\mathbf{65.88_{\pm 0.26}}$  &$65.83_{\pm 0.02}$  &$\mathbf{66.73_{\pm 0.08}}$  &$65.63_{\pm 0.94}$  &$\mathbf{65.97_{\pm 0.67}}$\\
                         & 60\%    &$62.40_{\pm 0.68}$  &$\mathbf{63.83_{\pm 0.68}}$  &$63.80_{\pm 0.06}$  &$\mathbf{64.67_{\pm 0.03}}$  &$62.15_{\pm 0.81}$  &$\mathbf{63.30_{\pm 0.65}}$\\
                         & 80\%    &$61.34_{\pm 0.60}$  &$\mathbf{62.75_{\pm 0.75}}$  &$62.84_{\pm 0.05}$  &$\mathbf{63.69_{\pm 0.05}}$  &$60.01_{\pm 0.91}$  &$\mathbf{61.70_{\pm 0.70}}$\\
                         & 100\%    &$59.83_{\pm 0.72}$  &$\mathbf{61.98_{\pm 0.46}}$  &$61.30_{\pm 0.06}$  &$\mathbf{62.94_{\pm 0.06}}$  &$58.82_{\pm 1.40}$  &$\mathbf{59.82_{\pm 0.99}}$\\
   \hline
    \multirow{5}{*}{PubMed} 
                         & 20\%    &$82.98_{\pm 0.08}$  &$\mathbf{83.02_{\pm 0.09}}$  &$71.19_{\pm 0.01}$  &$\mathbf{79.66_{\pm 0.26}}$  &$82.00_{\pm 0.12}$  &$\mathbf{82.55_{\pm 0.10}}$\\
                         & 40\%    &$81.13_{\pm 0.08}$  &$\mathbf{81.16_{\pm 0.06}}$  &$64.61_{\pm 0.02}$  &$\mathbf{76.43_{\pm 0.21}}$  &$80.01_{\pm 0.24}$  &$\mathbf{80.80_{\pm 0.20}}$\\
                         & 60\%    &$79.16_{\pm 0.08}$  &$\mathbf{79.24_{\pm 0.13}}$  &$60.64_{\pm 0.09}$  &$\mathbf{73.08_{\pm 0.42}}$  &$78.14_{\pm 0.18}$  &$\mathbf{79.54_{\pm 0.26}}$\\
                         & 80\%    &$77.88_{\pm 0.15}$  &$\mathbf{78.13_{\pm 0.20}}$  &$58.93_{\pm 0.07}$  &$\mathbf{70.50_{\pm 0.31}}$  &$76.90_{\pm 0.22}$  &$\mathbf{78.41_{\pm 0.13}}$\\
                         & 100\%    &$76.86_{\pm 0.18}$  &$\mathbf{77.04_{\pm 0.11}}$  &$57.45_{\pm 0.06}$  &$\mathbf{66.12_{\pm 0.89}}$  &$75.76_{\pm 0.19}$  &$\mathbf{77.18_{\pm 0.24}}$\\
    \hline
   \multirow{5}{*}{Polblogs}
                         & 20\%    &$89.37_{\pm 0.34}$  &$\mathbf{89.70_{\pm 1.41}}$  &$83.96_{\pm 0.76}$  &$\mathbf{89.26_{\pm 0.58}}$  &$90.38_{\pm 0.33}$  &$\mathbf{90.51_{\pm 0.43}}$\\
                         & 40\%    &$86.83_{\pm 0.31}$  &$\mathbf{88.29_{\pm 0.63}}$  &$75.97_{\pm 0.77}$  &$\mathbf{87.48_{\pm 0.18}}$  &$87.08_{\pm 0.83}$  &$\mathbf{88.20_{\pm 0.61}}$\\
                         & 60\%    &$85.50_{\pm 0.38}$  &$\mathbf{86.49_{\pm 0.70}}$  &$72.82_{\pm 0.81}$  &$\mathbf{86.26_{\pm 0.73}}$  &$86.47_{\pm 0.83}$  &$\mathbf{86.50_{\pm 0.84}}$\\
                         & 80\%    &$85.93_{\pm 0.30}$  &$\mathbf{86.81_{\pm 0.88}}$  &$64.33_{\pm 0.89}$  &$\mathbf{83.24_{\pm 0.81}}$  &$85.38_{\pm 1.12}$  &$\mathbf{85.91_{\pm 0.53}}$\\
                         & 100\%    &$85.48_{\pm 0.40}$  &$\mathbf{86.04_{\pm 0.39}}$  &$68.90_{\pm 0.39}$  &$\mathbf{85.31_{\pm 0.36}}$  &$84.56_{\pm 1.44}$  &$\mathbf{85.52_{\pm 0.78}}$\\
   \bottomrule
  \end{tabular}%
  \vspace{-5pt}
 \end{table*}

 \begin{table}
  \caption{ECE (\%) (with $M = 10$ bins) about cross-entropy loss and joint-cluster loss.}
  \vskip 0.05in
  \label{prediction calibration}
  \centering
  \begin{tabular}{ccc}
    \toprule
    Methods  & PubMed    & Arxiv \\
    \midrule
    GCN+CE   & $10.18\%$ & $1.28\%$ \\
    GCN+JC   & $\mathbf{6.82\%}$  & $\mathbf{0.47\%}$ \\
  \bottomrule
\end{tabular}
\end{table}

\paragraph{Prediction calibration.}
In order to showcase the efficacy of our framework in prediction calibration, we employ GCN as the encoder and evaluate the impact using Expected Calibration Error (ECE)~\citep{ECE}, which partition predictions into $M$ equally-spaced bins (similar to the reliability diagrams) and taking a weighted average of the bins’ accuracy/confidence difference~\citep{overcon}. We perform experiments on two citation networks(Pubmed and Arxiv), whose classes are the most(40) and least(3) respectively, to demonstrate the comprehensive nature of our method in prediction calibration. As shown in table~\ref{prediction calibration}, our framework provides superior prediction calibration than independent CE loss. Joint-cluster distribution learning produces more reliable classifications than independent decisions due to the reference signal of the clusters.

\section{Limitations and Future Work}\label{appendix_limit}
Although our framework achieves promising experimental justifications, it suffers from the computation inefficiency issue. 
Compared with the standard supervised learning, the joint-cluster distribution modeling expands a $c$-classes node classification task into a $c^{2}$-classes prediction problem. Consequently, we require the larger memory and more expensive time cost especially for the graph data with a large number of node classes. However, this computation challenge can be relieved by reformulating the $c^{2}$-classes prediction problem to a $2c$-classes setting, where the ground-truth probability values are described by the corresponding node or cluster labels. 

In the future work, we will explore the joint-cluster supervised learning on a broad range of potential applications, such as graph classification or link prediction. In addition, the correlation between samples is the biggest challenge in modeling real problems using probability theory, especially in graph data. We expect more studies and exploration on more intermediate factorizations between i.i.d and fully joint learning about the graph domain. We believe that the joint distribution learning will continue to be a promising research area.

\end{document}